\documentclass[10pt,twocolumn,letterpaper]{article}

\usepackage{cvpr}      %

\usepackage{graphicx}
\usepackage{amsmath}
\usepackage{amssymb}
\usepackage{booktabs}
\usepackage{algorithm,algorithmic}
\usepackage{colortbl}
\usepackage{graphicx}
\usepackage{tabularx}
\usepackage{amsmath}
\usepackage{amssymb}
\usepackage{booktabs}
\usepackage{url}
\usepackage{epsfig}
\usepackage{lipsum}
\usepackage{amsthm}
\usepackage{comment}
\usepackage{multirow}
\usepackage{subcaption}
\usepackage{microtype}
\usepackage{xspace}
\usepackage[dvipsnames]{xcolor}
\usepackage{enumitem}
\usepackage{tabularx}
\usepackage{graphbox}

\usepackage{amsmath,amsfonts,bm, bbm}

\def\eqref#1{equation~\ref{#1}}

\def\1{\bm{1}}

\def\rvu{{\mathbf{i}}}

\def\rvu{{\mathbf{u}}}

\def\vs{{\bm{s}}}

\def\vu{{\bm{u}}}

\def\vw{{\bm{w}}}
\def\vx{{\bm{x}}}
\def\vy{{\bm{y}}}

\def\mI{{\bm{I}}}

\DeclareMathAlphabet{\mathsfit}{\encodingdefault}{\sfdefault}{m}{sl}
\SetMathAlphabet{\mathsfit}{bold}{\encodingdefault}{\sfdefault}{bx}{n}

\def\gD{{\mathcal{D}}}

\def\gH{{\mathcal{H}}}

\def\gN{{\mathcal{N}}}

\newcommand{\E}{\mathbb{E}}

\newcommand\norm[1]{\left\lVert#1\right\rVert}

\newlength\myindent
\usepackage{amsmath}

\newcommand{\codecomment}[1]{\textcolor{gray}{#1}}

\usepackage{soul,xcolor}

\newcommand\redsout{\bgroup\markoverwith{\textcolor{red}{\rule[0.5ex]{2pt}{0.4pt}}}\ULon}

\newcommand{\our}{\texttt{DiffCollage}\xspace}
\newcommand{\ours}{\texttt{DiffCollage}\xspace}

\definecolor{cRUN}{RGB}{240, 162, 60}
\definecolor{cKICK}{RGB}{44, 153, 250}
\definecolor{cSKIP}{RGB}{138, 54, 224}
\definecolor{cBEND}{RGB}{55, 240, 131}
\definecolor{lightgray}{gray}{0.95}

\usepackage[export]{adjustbox}
\usepackage[pagebackref,breaklinks,colorlinks]{hyperref}
\usepackage{multirow}
\usepackage{ifthen}
\allowdisplaybreaks
\newboolean{arxiv}
\setboolean{arxiv}{false}

\usepackage[capitalize]{cleveref}
\crefname{section}{Sec.}{Secs.}
\Crefname{section}{Section}{Sections}
\Crefname{table}{Table}{Tables}
\crefname{table}{Tab.}{Tabs.}

\def\cvprPaperID{719} %
\def\confName{CVPR}
\def\confYear{2023}

\begin{document}

\title{DiffCollage: Parallel Generation of Large Content with Diffusion Models}

\author{
Qinsheng Zhang \and Jiaming Song \and Xun Huang \and Yongxin Chen \and Ming-Yu Liu 
\and{{\tt\small \{qzhang419,yongchen\}@gatech.edu}}\\ 
Georgia Institute of Technology
\and{{\tt\small \{jiamings,xunh,mingyul\}@nvidia.com}}\\
NVIDIA Corporation
}

\maketitle

\begin{abstract}
    We present \ours, a compositional diffusion model that can generate large content by leveraging diffusion models trained on generating pieces of the large content. Our approach is based on a factor graph representation where each factor node represents a portion of the content and a variable node represents their overlap. This representation allows us to aggregate intermediate outputs from diffusion models defined on individual nodes to generate content of arbitrary size and shape in parallel without resorting to an autoregressive generation procedure. We apply \ours to various tasks, including infinite image generation, panorama image generation, and long-duration text-guided motion generation. Extensive experimental results with a comparison to strong autoregressive baselines verify the effectiveness of our approach. \href{https://research.nvidia.com/labs/dir/diffcollage/}{project page}
\end{abstract}
\section{Introduction}\label{sec:intro}

\begin{figure*}[t!]
    \centering
    \adjincludegraphics[width=\linewidth,trim={{0cm} {13cm} {0cm} {0cm}},clip]{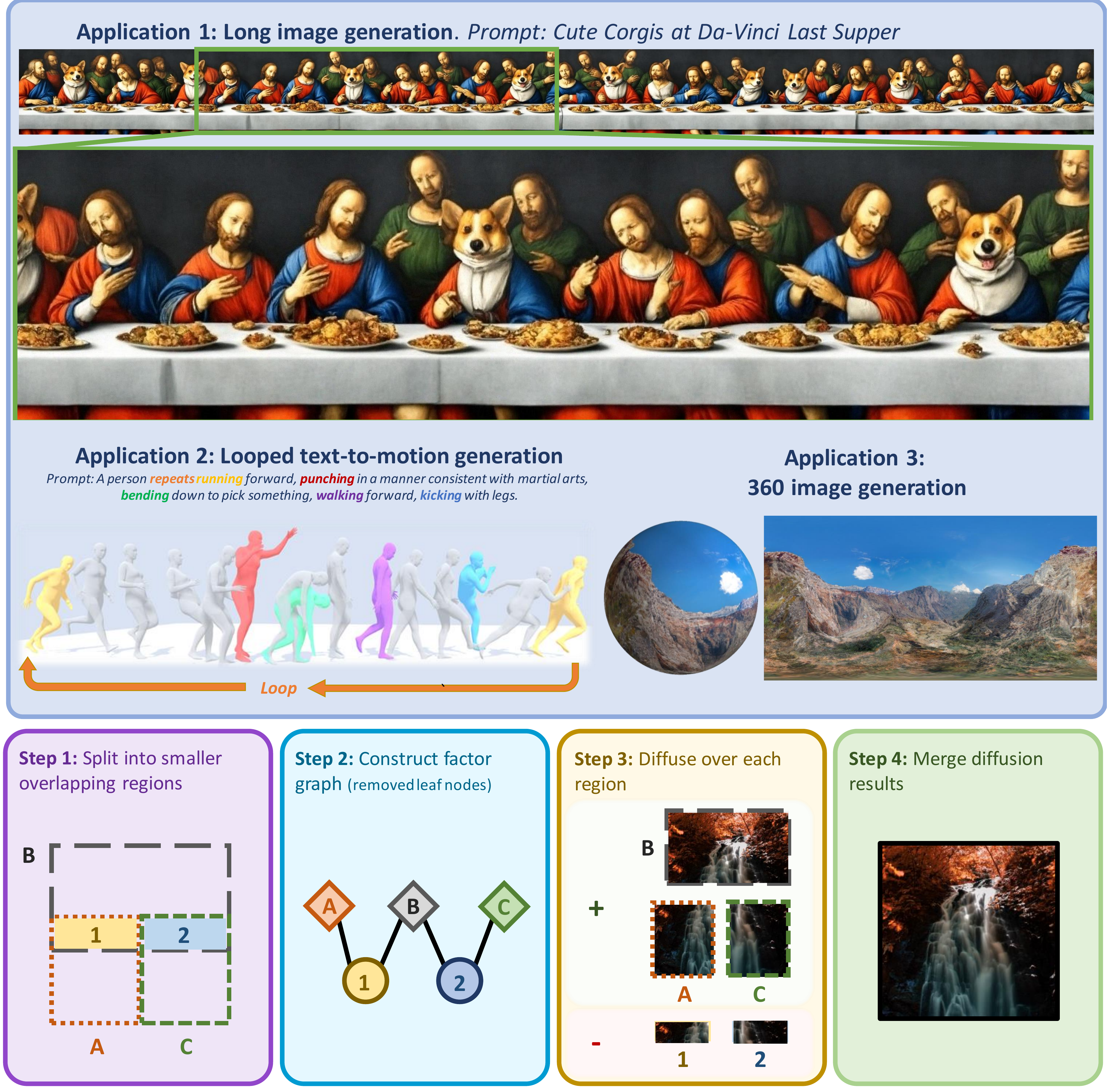}
    \captionof{figure}{
        \ours{}, a scalable probabilistic model that synthesizes large content, including long images, looped motions, and 360 images, with diffusion models only trained on pieces of the content.
    }
    \label{fig:teaser}
\end{figure*}

The success of diffusion models~\cite{ho2020denoising,SonSohKin2020} can largely be attributed to their scalability. With large-scale datasets and computing resources, practitioners can usually train high-capacity models that are able to produce high-fidelity images. 
The recent generative AI revolution led by large-scale text-to-image diffusion models is a great example~\cite{ramesh2022,saharia2022photorealistic,balaji2022ediffi}. The same procedure, collecting a large dataset and using it to train a large-scale model, has been applied to various problems and achieved great success~\cite{radford2021learning,brown2020language}.

In this paper, we are interested in extending the success of diffusion models to a wider class of data. We focus on applications where a large-scale dataset of the target content does not exist or is prohibitively expensive to collect, but individual pieces of the content are available in great quantities. 360-degree panorama images are such an example. While 360-degree panorama images are considered niche image content and only exist in small quantities, there are a large number of normal perspective images available on the Internet, each of which can be treated as a piece of a 360-degree panorama image. Another example is generating images of extreme aspect ratios, as shown in~\cref{fig:teaser}. Each of the extreme-aspect-ratio images can be considered as the stitching of multiple images with normal aspect ratios. For such applications, while we cannot afford to collect a large-scale dataset of the target content to train a diffusion model, we wish to synthesize high-quality target content with a diffusion model trained on smaller pieces that are readily available.

A popular solution to this class of problems is to first train a diffusion model on small pieces of the content and then generate the large content piece by piece in an autoregressive manner~\cite{saharia2022palette,esser2021taming}. 
However, such an autoregressive approach has three drawbacks. 
First, as pieces are generated sequentially, the later-generated pieces have no influence on the prior-generated ones. Such a sequential scheme could lead to sub-optimal results, especially when there is a circular structure in the data. For example, it is hard to enforce consistency between the start and end frames when generating looped videos autoregressively. 
Second, autoregressive methods may suffer from error accumulation since the model was conditioned on ground-truth data during training but is conditioned on its own prediction at test time.
Lastly, the time consumption of autoregressive generation increases linearly with the size of the data and could become prohibitive when generating very large content.

To address the large content generation problem, we propose \ours, a generic algorithm that synthesizes large content by merging the results generated by diffusion models trained on small pieces of the large content. 
Our approach is based on a factor graph formulation where a datum is modeled by a set of nodes and the edges connecting them. 
In our formulation, each node represents a contiguous portion of the large content, and the portions of content in neighboring nodes have a small overlap. 
Each node is associated with a small diffusion model and each piece affects the generation of the other pieces. Our method generates multiple pieces of content in parallel, which can greatly accelerate sampling when a large pool of computation is available.

We evaluate our approach on multiple large content generation tasks, including infinity image generation, long-duration text-to-motion with complex actions, content with unusual structures such as looped motion, and 360-degree images. 
Experiment results show that our approach outperforms existing approaches by a wide margin.

In summary, we make the following contributions.
\begin{itemize}[leftmargin=8pt]
    \setlength\itemsep{0em}
    \item We propose \ours, a scalable probabilistic model that synthesizes large content by merging results generated by diffusion models trained on pieces of the large content. It can synthesize large content efficiently by generating pieces in parallel. 
    \item \ours can work out-of-the-box when pre-trained diffusion models on different pieces are available.
    \item Extensive experimental results on benchmark datasets show the effectiveness and versatility of the proposed approaches on various tasks.
\end{itemize}

\section{Related Work}\label{sec:related}

\paragraph{Diffusion models }
Diffusion models~\cite{sohl2015deep,ho2020denoising,SonSohKin2020} have achieved great success in various problems, such as text-to-image generation~\cite{ramesh2022,saharia2022photorealistic,balaji2022ediffi}, time series modeling~\cite{tashiro2021csdi}, point cloud generation~\cite{zeng2022lion,ye2022first,zhou20213d}, natural language processing~\cite{li2022diffusion}, image editing~\cite{meng2021sdedit,kawar2022imagic,couairon2022diffedit,valevski2022unitune}, inpainting~\cite{lugmayr2022repaint,kawar2022denoising,choi2021ilvr,kawar2021snips}, and adversarial defense~\cite{nie2022DiffPure}. Recently, impressive progress has been made in improving its quality~\cite{saharia2022photorealistic,rombach2022high,ramesh2022,balaji2022ediffi}, controllability~\cite{ho2022classifier,meng2021sdedit,ruiz2022dreambooth,gal2022image,hertz2022prompt,kawar2022imagic}, and efficiency~\cite{song2020denoising,zhang2022fast,zhang2022gddim,jing2022subspace}. In this paper, we aim to enlarge the kind of data that diffusion models can generate.

\paragraph{Large content generation } Generating large content with generative models trained on small pieces of large content has been explored by prior works. 
One class of methods relies on latent variable models, \eg, GANs~\cite{goodfellow2014generative}, that map a global latent code and a spatial latent code to an output image. The global latent code represents the holistic appearance of the image and the spatial latent code is typically computed from a coordinate system. Some works~\cite{lin2019coco,skorokhodov2021aligning} generate different patches using the same global code and merge them to obtain the full image. A discriminator can be used to ensure the coherence of the full image. Instead of generating patches independently, some recent works generate the full image in one shot using architectures that guarantee translation equivariance, such as padding-free generators~\cite{lin2021infinitygan,ntavelis2022arbitrary,struski2022locogan} or implicit MLP-based generators~\cite{anokhin2021image,skorokhodov2021adversarial,chai2022any}.

Another popular approach to generating large content is to autoregressively apply ``outpainting'' to gradually enlarge the content. The outpainting could be implemented by a diffusion model~\cite{saharia2022palette,lugmayr2022repaint,chung2021come,kawar2022denoising,ho2022video,chung2022improving}, an autoregressive transformer~\cite{esser2021taming,wu2022nuwa,chen2022vector}, or a masked transformer~\cite{zhang2021m6,chen2022vector,chang2022maskgit}. 

\section{Preliminaries}

Diffusion models consist of two processes: a forward diffusion process and a reverse process.
The forward diffusion process progressively injects Gaussian noise into samples from the data distribution $q_0(\vu_0)$ and results in a family of noised data distributions $q_t(\vu_t)$. It can be shown that the distribution of $\vu_t$ conditioned on the clean data $\vu_0$ is also Gaussian: $q_{0t}(\vu_t | \vu_0) = \gN(\vu_0, \sigma_t^2 \mI)$. 
The standard deviation $\sigma_t$ monotonically increases with respect to the forward diffusion time $t$.
The reverse process is designed to iteratively remove the noise from the noised data to recover the clean data, which can be formulated as the following stochastic differential equation~(SDE)~\cite{karras2022elucidating,zhang2022gddim,SarSol19}
\begin{equation}~\label{eq:r-sde}
    d\vu = -(1+\eta^2)\dot{\sigma}_t \sigma_t \nabla_{\vu} \log q_t(\vu) dt + \eta \sqrt{ 2 \dot{\sigma}_t \sigma_t} d\vw,
\end{equation}
where $\nabla_\vu \log q_t(\vu)$ is the score function of a noised data distribution, $\vw_t$ is the standard Wiener process, and $\eta \geq 0$ determines the amount of random noise injected during the denoising process.
When $\eta=1$, \cref{eq:r-sde} is known as reverse-time SDE of the forward diffusion process\cite{SonSohKin2020,Anderson1982},
from which ancestral sampling and samplers based on Euler-Maruyama can be employed~\cite{SonSohKin2020,ho2020denoising}.
\cref{eq:r-sde} reduces to a probability flow ODE when $\eta=0$~\cite{SonSohKin2020}. %
In practice, the unknown score function $\nabla_\vu \log q_t(\vu)$ is estimated using a neural network $\vs_\theta(\vu, t)$ by minimizing a weighted sum of denoising autoencoder (score matching~\cite{Vin11}) objectives:
\begin{equation}\label{eq:dsm}
    \scalebox{1}{$\arg\min_{\theta} \E_{t,\vu_0} [
        \omega(t) \norm{
             \nabla_{\vu_t} \log q_{0t}(\vu_t | \vu_0) -
             \vs_\theta(\vu_t, t)
        }^2
    ],$}
\end{equation}
where $\omega(t)$ denotes a time-dependent weight.

\section{Diffusion Collage}\label{sec:body}

\ours is an algorithm that can generate large content in parallel using diffusion models trained on data consisting of portions of the large content. We first introduce the data representation and then discuss training and sampling. For simplicity, we derive the formulation for unconditional synthesis throughout this section; the formulation can be easily extended to conditional synthesis.

\subsection{Representation}

\begin{figure}
    \centering
     \includegraphics[width=\linewidth]{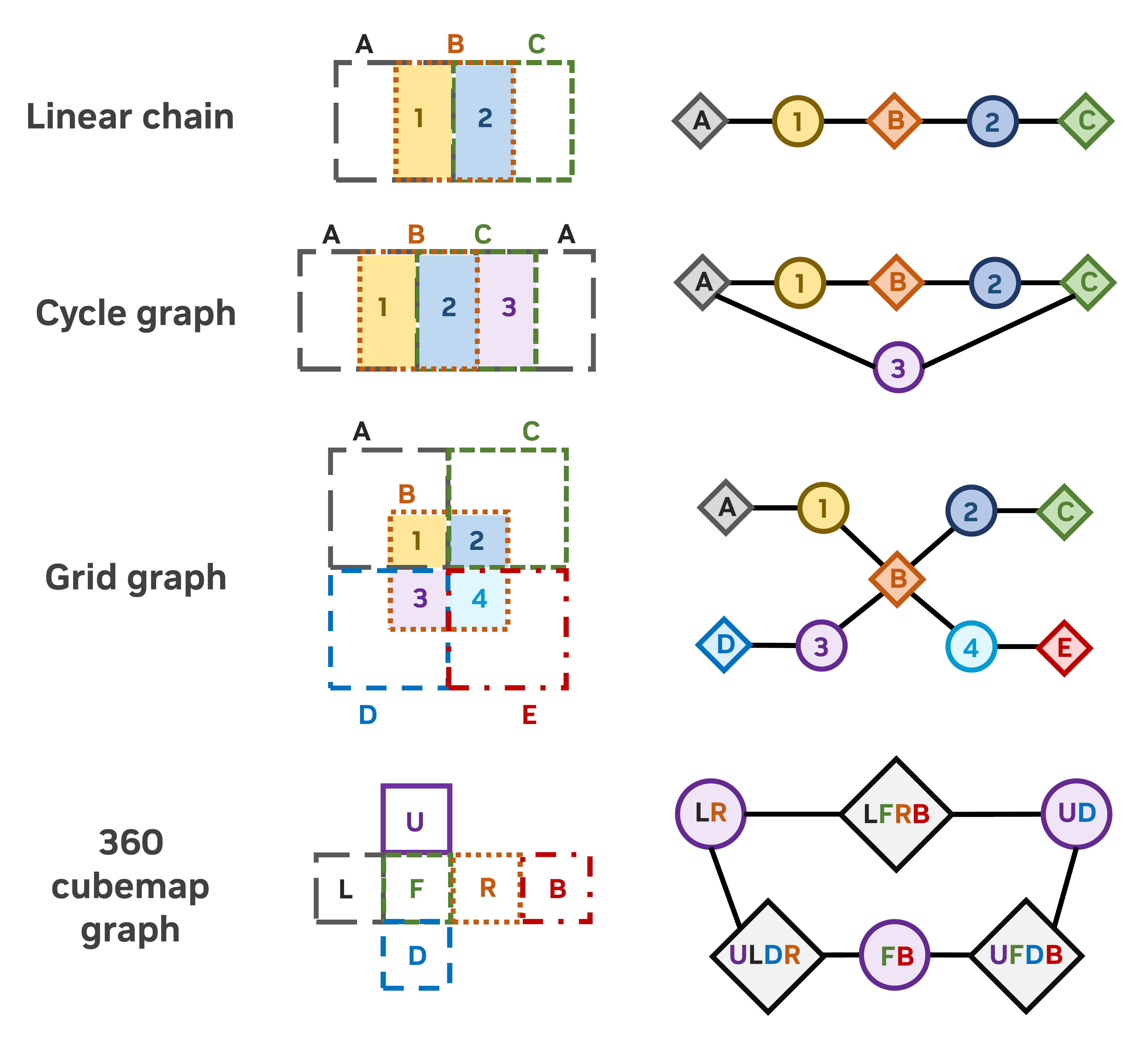}
     \caption{\textbf{Factor graphs for various applications.} From top to bottom: a linear chain for arbitrarily long sequences, a cycle graph for arbitrarily long loops, a grid graph for images of arbitrary height and width, and a complex factor graph for 360-degree panoramas.}
     \label{fig:example_graphs}
\end{figure}
\begin{figure}[t!]
    \centering
    \adjincludegraphics[width=\linewidth,trim={{0cm} {18cm} {12cm} {0cm}},clip]{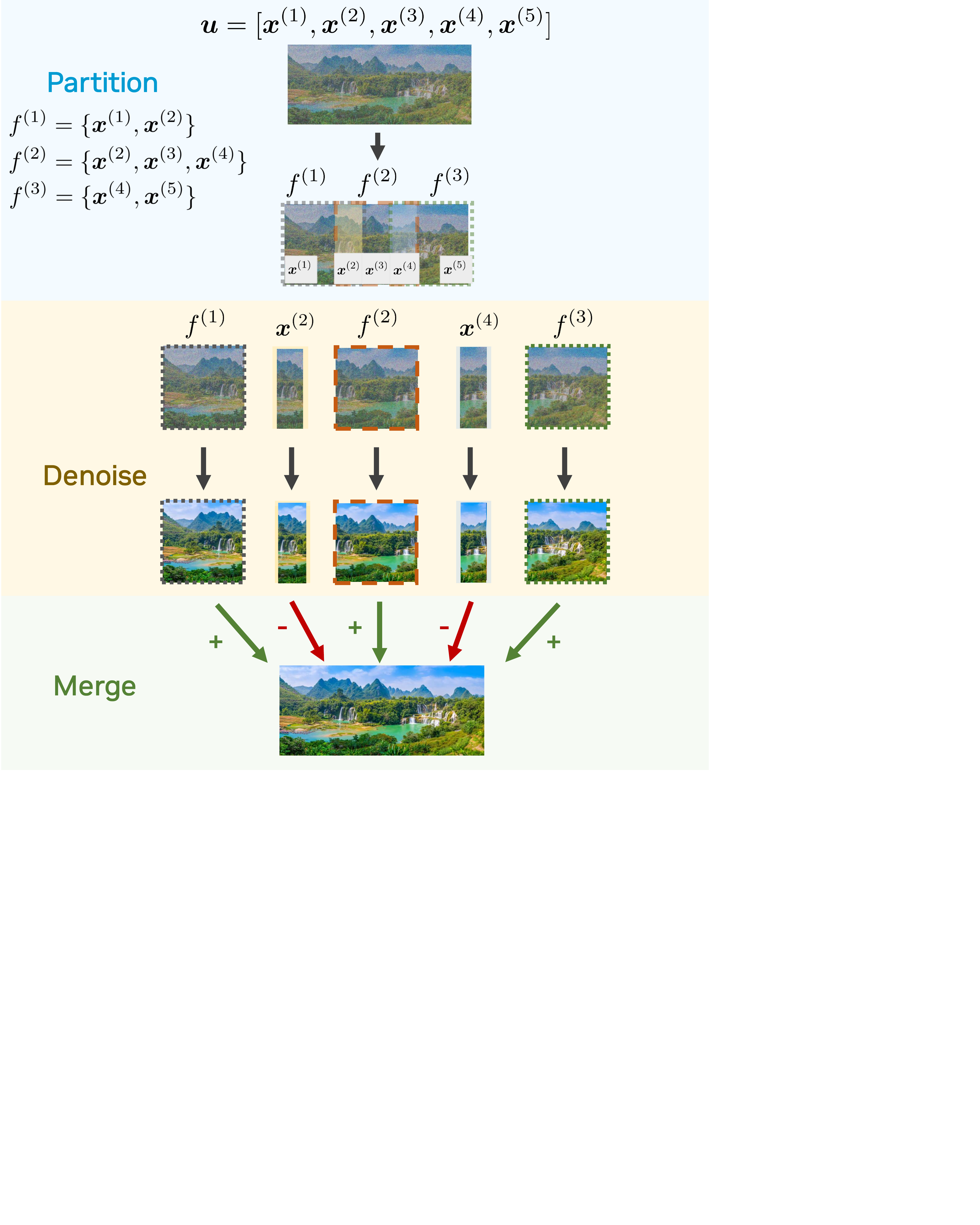}
    \caption{How \ours synthesize long images. To calculate the score for each denoising step on the long image, we split the input based on the factor graph into regions of factor nodes and variables. We obtain the scores of individual regions by using the individual diffusion models. We then merge the scores to compute the score of the target diffusion model on the long image. %
    }
    \label{fig:line-image-demo}
\end{figure}

\paragraph{A simple example}
A simple use case of \ours is to generate a long image by assembling diffusion models trained on shorter images. 
An autoregressive solution to this problem is to first generate an initial square image and then perform outpainting conditioned on a part of the previously generated image~\cite{saharia2022palette}, which results in a slightly larger output image. 
We denote this larger image as $\vu = [\vx^{(1)}, \vx^{(2)}, \vx^{(3)}]$ where $[\vx^{(1)}, \vx^{(2)}]$ is the initial image and $\vx^{(3)}$ is the outpainted region generated by the conditional model $\vx^{(3)} | \vx^{(2)}$.
Notably, this procedure makes a conditional independence assumption: conditioned on $\vx^{(2)}$, $\vx^{(1)}$ and $\vx^{(3)}$ are independent, \textit{i.e.}, $q(\vx^{(3)} | \vx^{(1)}, \vx^{(2)}) = q(\vx^{(3)} | \vx^{(2)})$. 
Therefore, the joint probability is
\begin{align}
   q(\vu) &= q(\vx^{(1)}, \vx^{(2)}, \vx^{(3)}) = q(\vx^{(1)}, \vx^{(2)}) q(\vx^{(3)} | \vx^{(2)}) \nonumber \\
    &= \frac{q(\vx^{(1)}, \vx^{(2)}) q(\vx^{(2)}, \vx^{(3)})}{q(\vx^{(2)})}\,. \label{eq:simple-joint}
\end{align}
The score function of $q(\vu)$ can be represented as a sum over the scores of smaller images
\begin{align}
  \nabla\log q(\vu) =& \ \nabla\log q(\vx^{(1)}, \vx^{(2)}) + \nabla\log q(\vx^{(2)}, \vx^{(3)})  \nonumber \\ & - \nabla\log q(\vx^{(2)})\,. \label{eq:simple-joint-score}
\end{align}

Each individual score can be estimated using a diffusion model trained on smaller images. Unlike the autoregressive method, which generates content sequentially, \ours can generate different pieces in parallel since all individual scores can be computed independently.

\paragraph{Generalization to arbitrary factor graphs} Now, we generalize the above example to more complex scenarios. 
For a joint variable $\vu = [\vx^{(1)}, \vx^{(2)}, \ldots, \vx^{(n)}]$, a factor graph \cite{KolFri09} is a bipartite graph connecting variable nodes $\{\vx^{(i)}\}_{i=1}^{n}$ and factor nodes $\{f^{(j)}\}_{j=1}^{m}$, where $f^{(j)}\subseteq \{\vx^{(1)}, \vx^{(2)}, \ldots, \vx^{(n)}\}$.
An undirected edge between $\vx^{(i)}$ and $f^{(j)}$ exists if and only if $\vx^{(i)} \in f^{(j)}$. 
In the above example, there are two factors $f^{(1)} = \{\vx^{(1)}, \vx^{(2)}\}$ and $f^{(2)} = \{\vx^{(2)}, \vx^{(3)}\}$. 
Given a factor graph that represents the factorization of the joint distribution $q(\vu)$\footnote{In order words, the joint distribution can be written as a product of functions, each of which is a function of a single factor.}, \ours approximates the distribution as follows: 
\begin{align}
    p(\vu) := \frac{\prod_{j=1}^{m} q(f^{(j)})}{\prod_{i=1}^{n} q(\vx^{(i)})^{d_i - 1}}\,, \label{eq:bethe}
\end{align}
where $d_i$ is the degree of the variable node $\vx^{(i)}$. It is easy to verify that \cref{eq:bethe} reduces to \cref{eq:simple-joint} in the simple case since the nodes for $\vx^{(1)}$ and $\vx^{(3)}$ have a degree of one (connected to $f^{(1)}$ and $f^{(2)}$ respectively) and the node for $\vx^{(2)}$ has a degree of two (connected to both $f^{(1)}$ and $f^{(2)}$). Similar to \cref{eq:simple-joint-score}, we can approximate the score of $q(\vu)$ by adding the scores over factor nodes (\textit{i.e.}, $q(f^{(j)})$) and subtracting the scores over non-leaf variable nodes (\textit{i.e.}, $q(\vx^{(i)})$)
\everymath{\displaystyle}
\begin{align}
    \scalebox{0.87}{$\nabla\log p(\vu) := \sum_{j=1}^{m} \nabla\log q(f^{(j)}) + \sum_{i=1}^{n}(1 - d_i) \nabla\log q(\vx^{(i)})\,.$}\label{eq:bethe-score}
\end{align}
In fact, \cref{eq:bethe} is also known in the probabilistic graphical model literature as the seminal \textit{Bethe approximation}, which approximates the joint distribution $q(\vu)$ by its marginals defined over factor and variable nodes~\cite{YedFreWei01,YedFreWei05}. The approximation is exact, \textit{i.e.}, $p(\vu) = q(\vu)$, when the factor graph is an acyclic graph. For a general graph with cycles, the Bethe approximation is widely used in practice and obtains good performance~\cite{singh2020inference,KolFri09}. More discussions and justification on Bethe approximation are in \ifthenelse{\boolean{arxiv}}{\cref{sec:appendix}}{the supplementary material}.

\begin{figure*}
    \centering
    \includegraphics[width=1.0\linewidth]{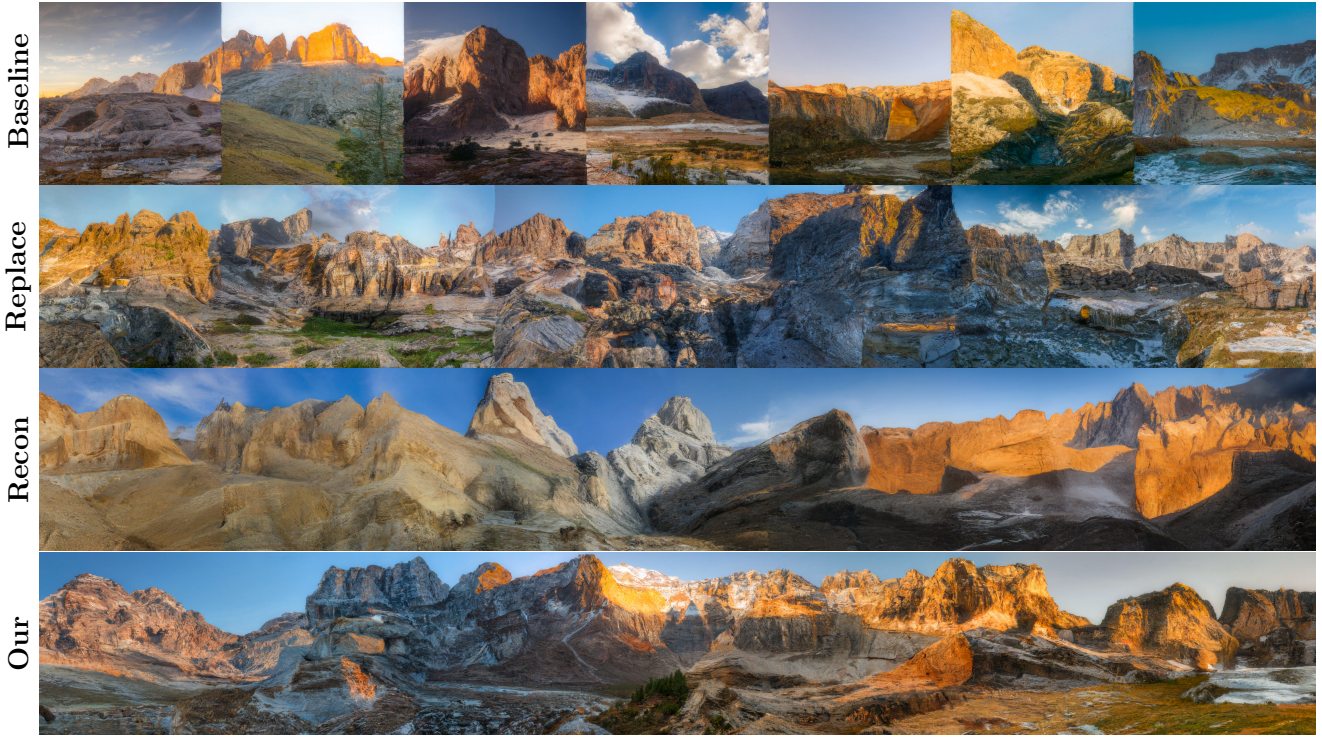}
    \caption{
        Long images generated by various approaches that only use diffusion models trained on smaller square images. 
        For autoregressive approaches~(Replace and Recon), we first generate the image in the middle then outpaint towards left and right.
        Replace and Recon introduce discontinuity artifacts while \ours can generate high-fidelity images in parallel.
    }
    \label{fig:method_comp}
\end{figure*}
In practice, factor graphs are general enough to cover contents of arbitrary size and shape, such as those in \cref{fig:example_graphs}:
\begin{itemize}[leftmargin=8pt]
    \setlength\itemsep{0em}
    \item \textbf{An arbitrarily long sequence}: The factor graph is a linear chain in which each factor is connected to two variables, and each variable is connected to two factors (except for leaf variables). We show a detailed characterization in \cref{fig:line-image-demo}.
    \item \textbf{An arbitrarily long sequence with a loop}: Similar to the linear chain but with a variable node connecting the head and tail factor nodes.
    \item \textbf{An image of arbitrary height and width}: Here, each factor is an image patch that overlaps with 4 other factors at the 4 corners. Each overlapping region is a variable. Thus, each factor node is connected to 4 variables, and each variables node is connected to 2 factors (except for edge cases).
    \item \textbf{A 360-degree image represented as a cubemap}: A cube consists of 6 faces: \textbf{F}ront, \textbf{B}ack, \textbf{L}eft, \textbf{R}ight, \textbf{U}p, \textbf{D}own. 
    There are three cycles (LFRB, ULDR, UFDB) can be modeled via the cycle graph, and these cycles overlap one another by two faces. Intuitively, we can treat these cycles as factors and faces as variables. We list more details in \ifthenelse{\boolean{arxiv}}{\cref{sec:appendix}}{the supplementary material}.
\end{itemize}

\subsection{Training and Sampling}

\paragraph{Training}
\ours is trained to estimate the score of noised data distributions $q_t(\vu)$.
Similar to the Bethe approximation (\cref{eq:bethe,eq:bethe-score}) for clean data, we factorize the score of the time-dependent noised data distributions:
\begin{align}\label{eq:dc-score}
    \scalebox{1}{$\nabla \log p_\theta(\vu,t) = $} \
    & \scalebox{1}{$ \sum_{j=1}^m \nabla \log p_\theta(f^{(j)},t)$}  \nonumber \\
    & \scalebox{1}{$+ \sum_{i=1}^n (1 - d_i) \nabla \log p_\theta(\vx^{(i)},t).$}
\end{align}

To close the gap between our learned model and the Bethe approximation in~\cref{eq:bethe}, we optimize $\theta$ by performing denoising score matching between the marginal scores of real data $\{ q(\vx^{(i)},t), q(f^{(j)},t) \}$ and learned marginal scores $\{ p_\theta(\vx^{(i)}, t), p_\theta(f^{(j)}, t ) \}$ (following \cref{eq:dsm}). This can be done by learning a diffusion model for each marginal distribution of real data; we list the detailed algorithm for training in the supplementary material.

It should be noted that even though we aim to approximate one joint distribution, learning a diffusion model for one marginal distribution is independent of learning other marginals. 
With such independence, diffusion models on different marginals can be learned in parallel.
Practically, diffusion models on different marginals can be amortized where we employ one shared diffusion model with conditional signals $\vy[f^{(j)}]$ from factor node $f^{(j)}$ and $\vy[i]$ from variable node $\vx^{(i)}$ to learn various marginals.

\paragraph{Sampling} After training the diffusion models for each marginal, the score model of \ours for $p_\theta(\vu, t)$ is simply obtained via \cref{eq:dc-score}, and it is a diffusion model with a specific score approximation. Thus \ours is sampler-agnostic, and  
we can leverage existing solvers for~\cref{eq:r-sde} to generate samples with the approximated score in~\cref{eq:dc-score}, such as DDIM~\cite{song2020denoising}, DEIS~\cite{zhang2022fast}, DPM-Solver~\cite{lu2022dpm} and gDDIM~\cite{zhang2022gddim}, all without any modifications. We emphasize that diffuson models on various marginals can be evaluated at the same time and generate different pieces of data $\{f^{(j)}, \vx^{(i)}\}$ in parallel, unlike the conventional autoregressive approaches, so with advanced samplers, the number of iterations taken by \ours{} could be much less than that of an autoregressive model.

\section{Experiments}\label{sec:expr}
Here, we present quantitative and qualitative results to show the effectiveness and efficiency of \ours. 
We perform experiments on various generation tasks, such as infinite image generation (\cref{sec:exp-img-generation}), arbitrary-sized image translation~(\cref{sec:exp-image-translation}), motion synthesis (\cref{sec:exp:t2m-generation}), and 360-degree panorama generation (\cref{sec:exp-complex-graph}).

\subsection{Infinite image generation}\label{sec:exp-img-generation}

We first evaluate \ours in the infinite image generation task~\cite{skorokhodov2021aligning} where the goal is to generate images extended to infinity horizontally. We employ a linear chain as shown in~\cref{fig:example_graphs} and use {\bf the same} score network for all factor nodes and variable nodes since the marginal image distribution is shift-invariant. 

We finetune a pre-trained GLIDE model~\cite{Nichol2021a}, which is a two-stage diffusion model consisting of a $64\times 64$ square generator and one $64 \to 256$ upsampler on an internal landscape dataset. 
We additionally finetune a pre-trained eDiff-I~\cite{balaji2022ediffi} $256 \to 1024$ upsampler.
Combining them together, we have a score model for individual nodes that can generate images of resolution up to $1024\times 1024$. 
To control the style of the output~\cite{balaji2022ediffi,ramesh2022}, the base diffusion model is conditioned on CLIP~\cite{radford2021learning} image embeddings.

Other diffusion-based approaches tackle this problem by performing outpainting autoregressively~\cite{saharia2022palette}.
Specifically, it generates the first image using a standard diffusion model, then extends the image through repeated application of outpainting toward left and right.
The outpainting problem can be treated as an inpainting problem with 50\% of the content masked out.
While there exist diffusion models specifically trained for inpainting~\cite{saharia2022palette},
we only perform comparisons with other generic methods that work on any pretrained diffusion models. 
We compare \ours with two inpainting approaches.
The first one is the ``\textbf{replacement}'' approach, where we constantly replace part of intermediate predictions with known pixels~\cite{lugmayr2022repaint,kawar2022denoising,choi2021ilvr}.
The second is the ``\textbf{reconstruction}'' approach, which uses the gradient of the reconstruction loss on known pixels to correct the unconditional samples~\cite{ho2022video,chung2022improving,ryu2022pyramidal}. This approach is slightly more computationally expensive since it needs to compute the gradient of the reconstruction loss w.r.t the intermediate predictions. We discuss more details in the supplementary.

To compare the generation quality of generated panorama images, we propose \textit{FID Plus~(FID+)}. 
We first generate $50$k panorama images with spatial ratio $W/H=6$ and randomly crop one square image $H\times H$ for each image. FID+ is the Frechet inception distance (FID,~\cite{heusel2017gans}) of the $50$k randomly cropped images. 
We also include a \textbf{baseline} that naively concatenates independently generated images of $H \times H$ into a long image. 
Although each generated image is realistic, this approach has a bad FID+ because randomly cropped images may contain clear boundaries.

\begin{figure}
    \centering
    \includegraphics[width=0.495\linewidth, height=0.495\linewidth]{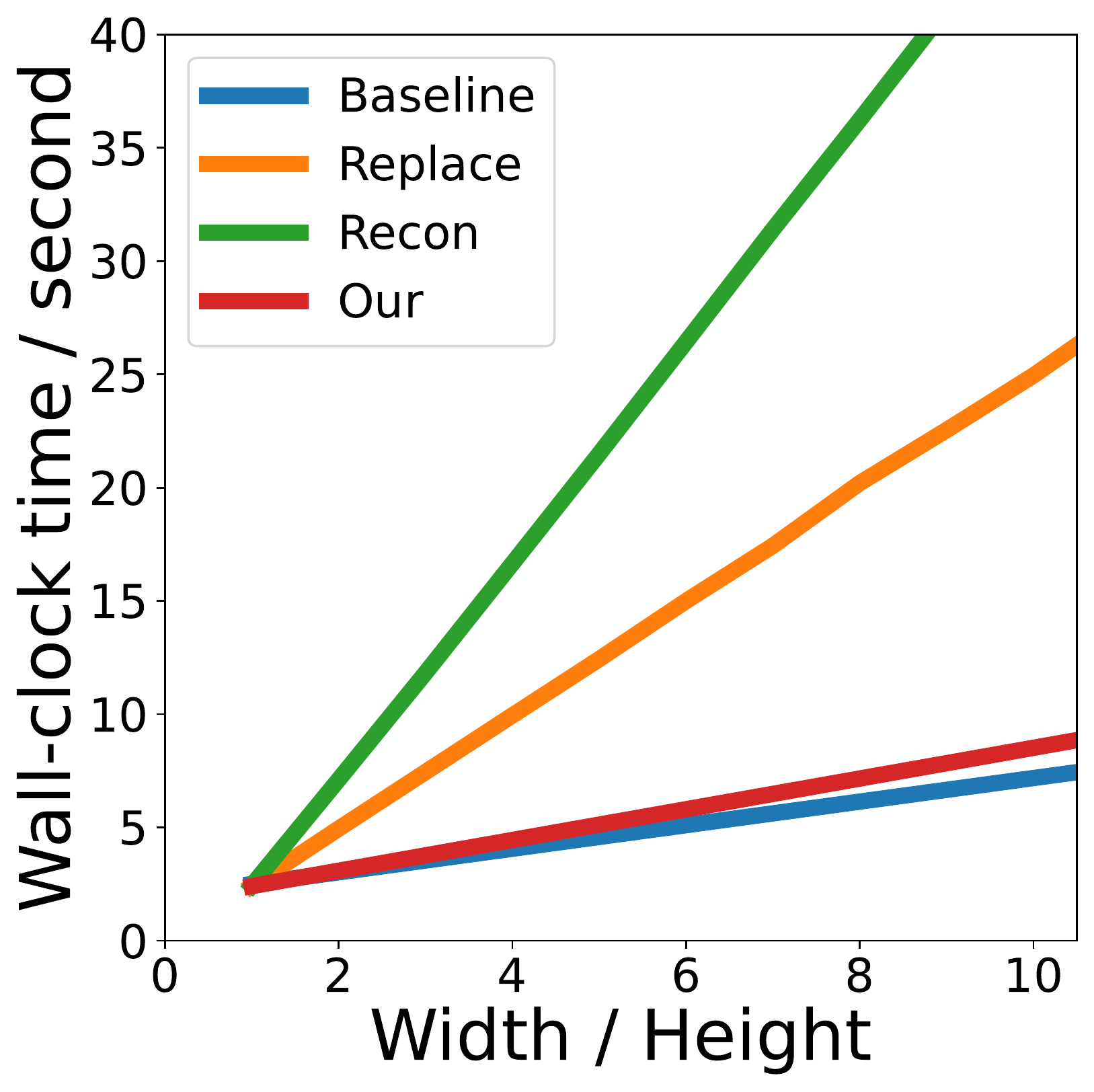}
    \includegraphics[width=0.495\linewidth, height=0.495\linewidth]{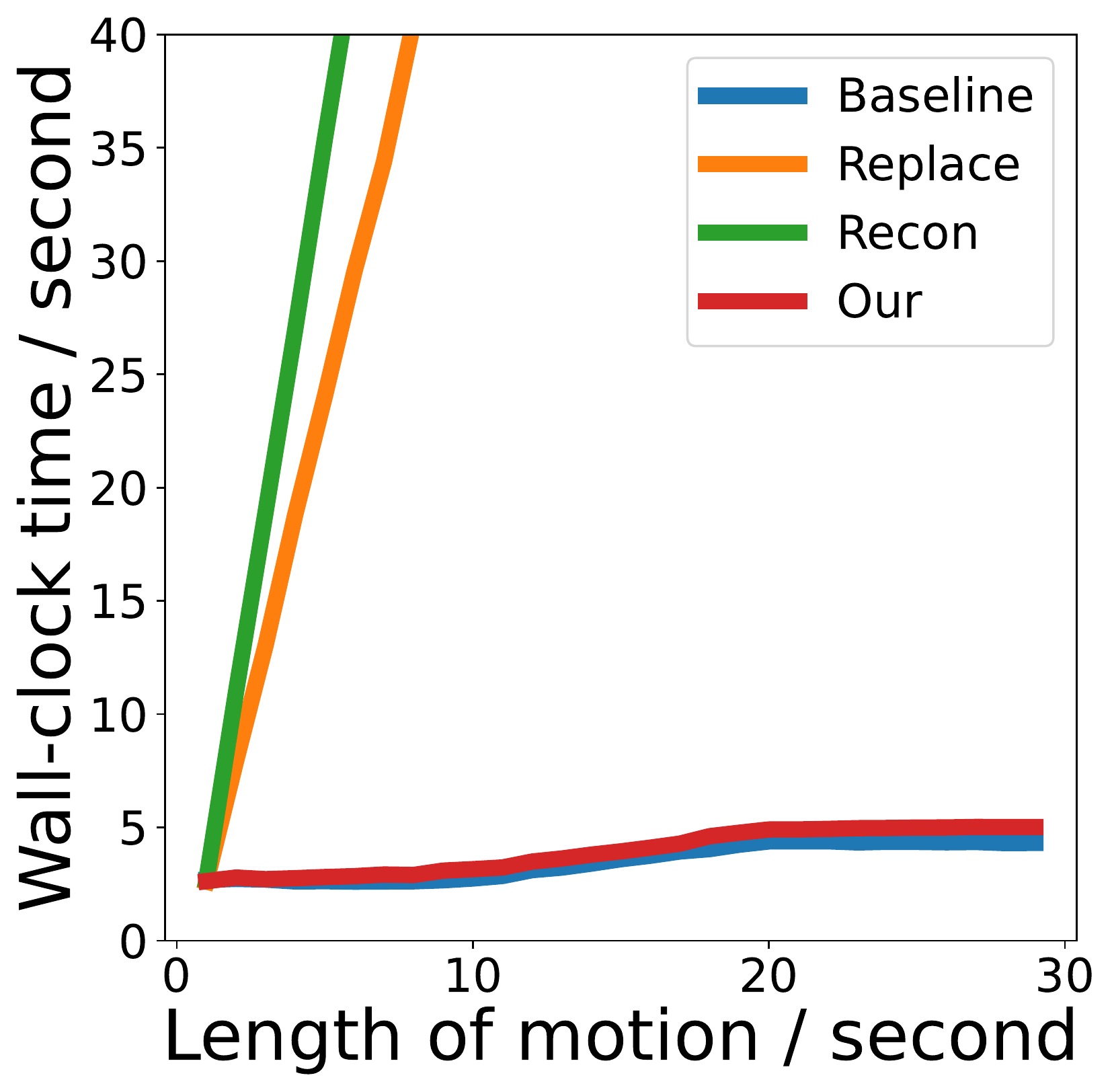}
    \caption{
        Wall-clock time for generating images with various lengths~(Left) and motion sequences with various durations~(Right). 
    }
    \label{fig:panorama-dm-t}
\end{figure}

As shown in~\cref{tab:panorama-dm}, \our{} outperforms other approaches in terms of sample quality evaluated by FID+. 
In \cref{fig:method_comp}, we show that the sample quality of autoregressive approaches deteriorates as the image grows due to error accumulation, while \our{} does not have this issue. %
\cref{fig:panorama-dm-t} compares the latency of generating one panorama image with different image sizes. Thanks to its parallelization, \our{} is about $ H / 2W $ times and $H/W$ times faster than replacement and reconstruction methods respectively when generating one $H\times W$ image with $H \geq 2W$. We further apply \ours{} to eDiff-I~\cite{balaji2022ediffi}, a recent large-scale text-to-image model in~\cref{fig:teaser}, to generate a wide image from the text prompt \textit{``Cute Corgis at Da-Vinci Last Supper''}, which demonstrates the general applicability of \ours to arbitrary diffusion models. 

\begin{figure*}[t!]
    \centering
    \includegraphics[width=\textwidth]{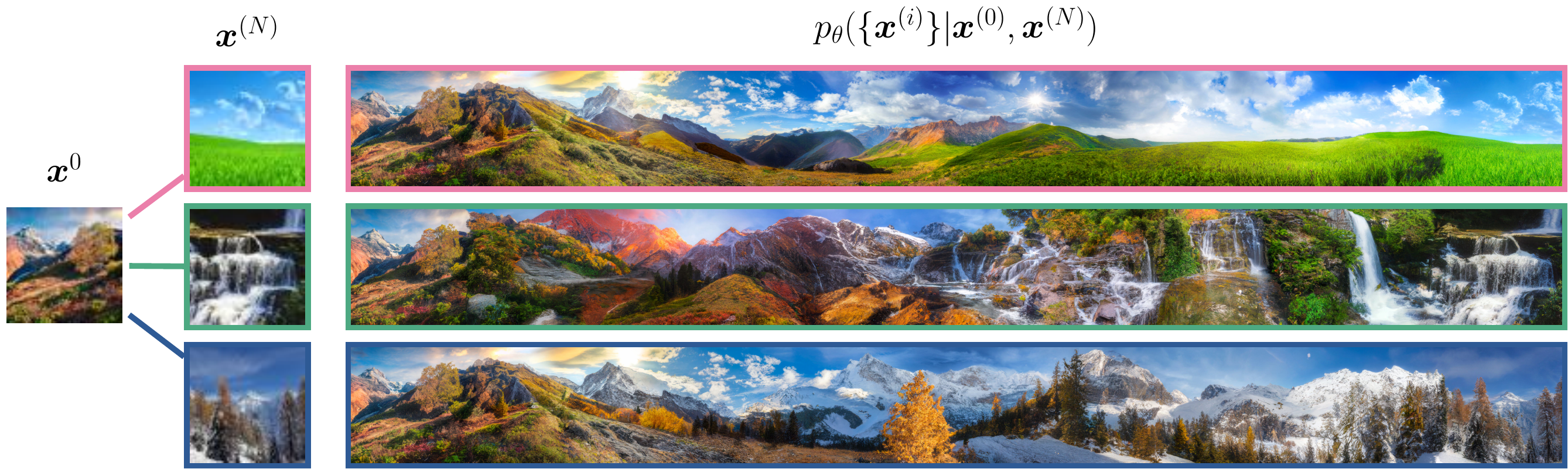}
    \caption{
        \textbf{Connecting real images.} 
        Given a pair of $64\times 64$ real images $\vx^{(0)}$ and $\vx^{(N)}$, \ours can generate a $1024\times 10752$ image that transitions naturally from $\vx^{(0)}$ into $\vx^{(N)}$.
    }
    \label{fig:connect}
\end{figure*}

In addition, the score models for different nodes can be conditioned on different control signals.
We illustrate this point by connecting any two landscape images of different styles. 
This is a challenging inpainting task where only pixels at two ends are given. 
The score models for intermediate nodes are conditioned on interpolated CLIP embeddings.
As shown in~\cref{fig:connect}, we are able to generate a long image that transitions from one style to another totally different one.

We compare \our{} with other methods specifically designed for long image generation tasks on LHQ~\cite{skorokhodov2021aligning} and LSUN Tower~\cite{yu2015lsun}, following the setting in Skorokhodov~\textit{et al.}~\cite{skorokhodov2021aligning}. We evaluate standard FID over images of size $H \times H$, as well as FID+ in \cref{tab:panorama-other}. As shown in the table, even though our approach is never trained on long image generation, it achieves competitive results compared with methods that are tailored to the task and require problem-specific networks for the image dataset.
\begin{table}[]
    \centering
    \scalebox{0.9}{
        \begin{tabular}{ccccc}
            \toprule
             Algo        & Parallel & Gradients & FID+~$\downarrow$ & Time~$\downarrow$\\
             \hline
             \cellcolor{lightgray} Baseline        &  \cellcolor{lightgray}-  & \cellcolor{lightgray}- & \cellcolor{lightgray}24.15 & \cellcolor{lightgray} 5.61 \\
             Replacement     &  {\color{red!70!black} No}  & {\color{green!50!black} Not required} & 10.25 & 14.99\\
             Reconstruction    &  {\color{red!70!black} No}  & {\color{red!70!black} Required} & 8.97 & 26.43\\
             Ours        & {\color{green!50!black} Yes}  & {\color{green!50!black} Not required}  & \textbf{4.54} & \textbf{6.47}\\
             \bottomrule
        \end{tabular}
        }
    \caption{Comaprison among diffusion-based methods for infinite image generation on an internal landscape dataset. Our method achieves the best image quality while also being the fastest since we can compute individual scores in parallel and do not require backpropagating through the diffusion model to obtain gradients.}
    \label{tab:panorama-dm}
\end{table}
\begin{table}[!tp]
    \centering
    \scalebox{1}{
    \begin{tabular}{cclcl}
    \toprule
\multirow{2}{*}{Dataset} & \multicolumn{2}{c}{LHQ $256^2$}         & \multicolumn{2}{c}{Tower $256^2$}       \\
                         & \multicolumn{1}{l}{FID} & FID+  & \multicolumn{1}{l}{FID} & FID+  \\
\midrule
\cellcolor{lightgray}VQGAN~\cite{esser2021taming}                   & \cellcolor{lightgray}58.27                   & \cellcolor{lightgray}62.12 & \cellcolor{lightgray}45.18                   & \cellcolor{lightgray}47.32 \\
\cellcolor{lightgray}ALIS~\cite{skorokhodov2021aligning}                     & \cellcolor{lightgray}12.60                   & \cellcolor{lightgray}14.27 & \cellcolor{lightgray}11.85                   & \cellcolor{lightgray}15.27 \\
Replacement              & 6.28                    & 28.94 & 7.15               &       30.19\\
Reconstruction           & 6.28                    & 18.37 & 7.15               &       19.56\\
Ours                     & 6.28                    & \textbf{16.43} & 7.15               &       \textbf{13.27}\\
\bottomrule
\end{tabular}
}
    \caption{Comparison against methods specifically designed for infinity image generation~(Dark-colored rows). Our approach achieves higher quality despite being more general.}
    \label{tab:panorama-other}
\end{table}

\begin{figure}[t!]
    \centering
    \setlength\tabcolsep{0.01cm}
    \begin{tabular}{lcc}

    \raisebox{0.7cm}[0pt][0pt]{\rotatebox[origin=c]{90}{Masked}} \hspace{1em}&\includegraphics[width=0.38\linewidth]{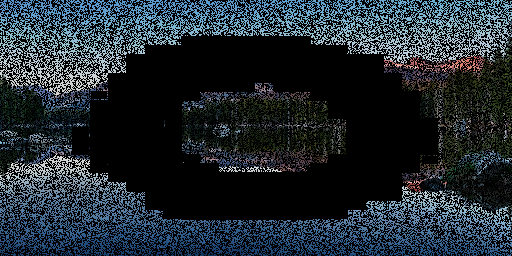}
    &\includegraphics[width=0.57\linewidth]{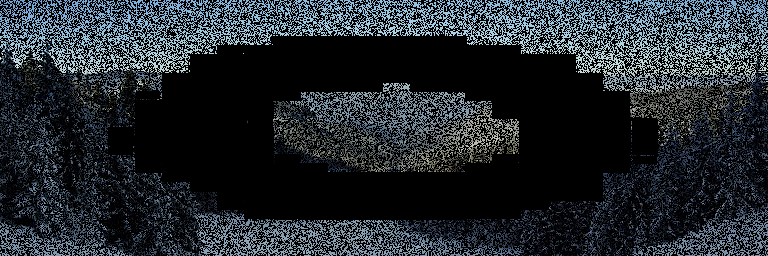} \\

    \raisebox{0.7cm}[0pt][0pt]{\rotatebox[origin=c]{90}{Recon}}&\includegraphics[width=0.38\linewidth]{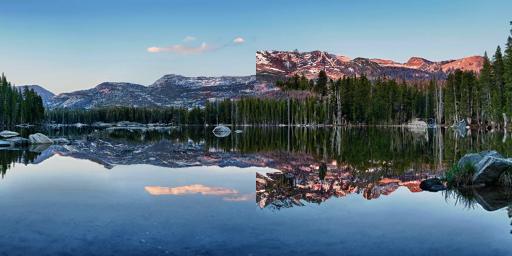}
    &\includegraphics[width=0.57\linewidth]{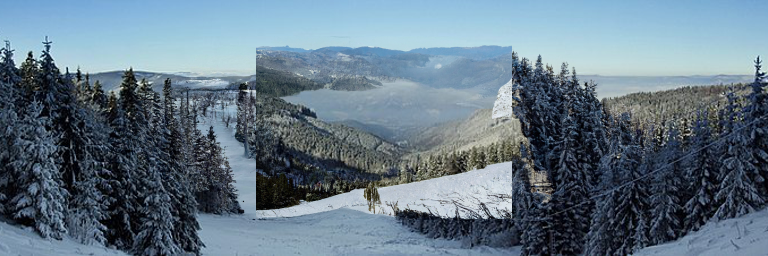} \\

    \raisebox{0.7cm}[0pt][0pt]{\rotatebox[origin=c]{90}{Ours}}&\includegraphics[width=0.38\linewidth]{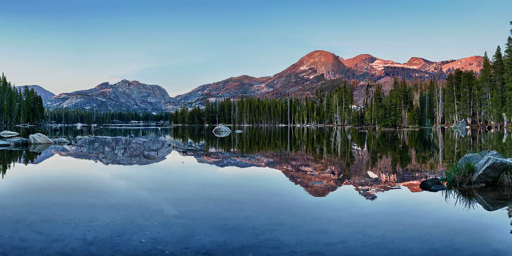}
    &\includegraphics[width=0.57\linewidth]{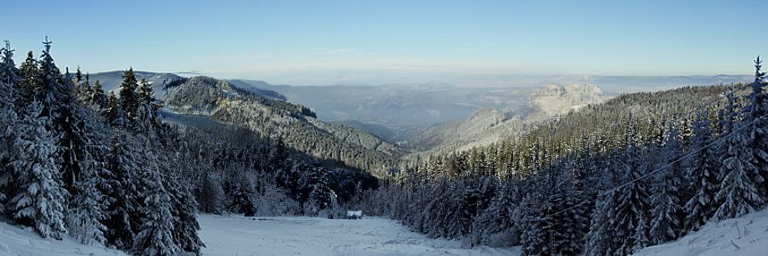}
    \end{tabular}
    \caption{
        \textbf{Inpainting on non-square images.} The first row contains two masked images. The second row contains the inpainting results by splitting the input non-square images into a set of square images. In the left/right example, the input image is split into two/three square images.  The Recon results in apparent boundary artifacts. The third row contains the inpainting results with \our{}.
    }
    \label{fig:panorama-inpainting}
\end{figure}
\subsection{Arbitrary-sized image translation} \label{sec:exp-image-translation}
Our method can be applied to various image translation tasks where the size of the input image is different from what the diffusion model is trained on. We use \ours to aggregate the scores of individual nodes and the score of each node can be estimated using methods that are developed for standard diffusion models,
such as replacement~\cite{lugmayr2022repaint,chung2021come,kawar2022denoising} or reconstruction methods~\cite{ho2022video,chung2022improving} for inpainting, and
SDEdit~\cite{meng2021sdedit} for stroke-based image synthesis. 
To achieve these with existing methods, one could also split the large image into several smaller ones and apply the conditional generation methods independently. However, this fails to model the interactions between the split images, resulting in discontinuities in the final image; we illustrate this in~\cref{fig:panorama-inpainting} for the task of inpainting from sparse pixels.
In contrast, \our{} can capture global information and faithfully recover images from given sparse pixels.
We include more image translation results in \ifthenelse{\boolean{arxiv}}{\cref{sec:appendix}}{the supplementary}.

\begin{figure*}[t!]
    \centering
    \adjincludegraphics[width=\textwidth,trim={0 {.84\height} 0 {.0\height}},clip]{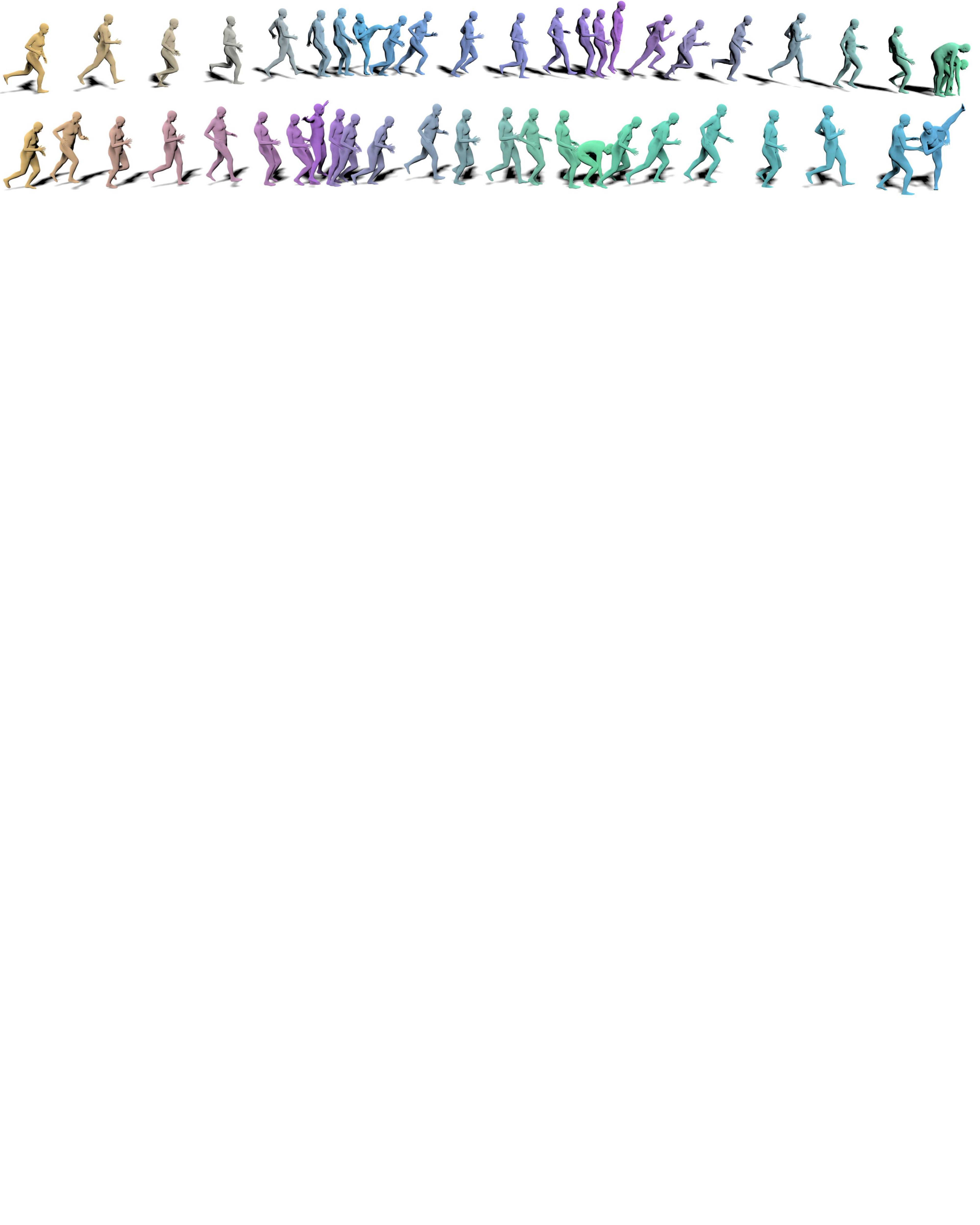}
    \caption{
        Complex motions synthesis. 
        Though the pre-trained motion diffusion model~\cite{tevet2022human} can only generate simple motions with one or two actions, \ours can extend it to synthesize long sequences with an arbitrary number of actions. Prompts:
        (Top) \textit{A person \textcolor{cRUN}{runs} forward, 
        then \textcolor{cKICK}{kicks his legs},
        then \textcolor{cSKIP}{skips rope},
        then \textcolor{cBEND}{bends down} to pick something up off the ground}.
        (Bottom) \textit{A person \textcolor{cRUN}{runs} forward, 
        then \textcolor{cSKIP}{skips rope},
        then \textcolor{cBEND}{bends down} to pick something up off the ground,
        then \textcolor{cKICK}{kicks his legs}}.
    }
    \label{fig:long-motion}
\end{figure*}

\subsection{Text-to-motion generation}\label{sec:exp:t2m-generation}
Given a text description of the desired motion, the goal of this task is to synthesize a motion sequence corresponding to the description.
In this section, we evaluate our approach on the popular benchmark HumanML3D~\cite{Guo_2022_CVPR,AMASS:2019}, using a pre-trained motion diffusion model~\cite{tevet2022human}.
The pre-trained diffusion model was trained with sequences of various lengths. As a result, it can use be used as the score estimator for both factor nodes and variable nodes in our formulation directly. %

High-fidelity generated samples are expected to follow basic rules of physics and behave similarly to realistic human motions. Specifically, we adopt the set of metrics from Guo~\textit{et al.}~\cite{Guo_2022_CVPR}, 
including \textit{R-precision} and \textit{Multimodal-Distance} that quantify the alignment between generated samples and the given prompt, 
\textit{FID} that measures the distance between the distribution of ground truth motions and generated motions, and \textit{Diversity} that measures the variability in samples generated by our methods.
    
\textbf{Long-duration motion generation} 
In HumanML3D, the average motion length is $7.1$s, and the maximum duration is $10$s. Our goal is to generate high-fidelity motion sequences that are much longer than what we have in the training data. To achieve this, we use a linear chain graph similar to the one used in infinite image generation. To evaluate our method, we generate a $24$s motion for each text and randomly crop generated sequences, analogous to FID+ for images.

We compare our approach with several methods, including naively denoising a long sequence (Baseline) and autoregressive generation with replacement and reconstruction methods, respectively. The results in~\cref{tab:motion-seq-perf} show that~\our{} outperforms other approaches in all evaluated metrics by a notable amount.

\textbf{Compositing multiple actions}
The existing human motion generative model can only synthesize simple motions since there are only one or two actions for one motion sequence in the training dataset. 
With \our{}, we can augment the simple motion diffusion model with the ability to synthesize complex actions. We use the desired text prompts for the conditions of factors $\vy[f_j]$, and the unconditional null token for that of variables $\vy[i]$.
As shown in~\cref{fig:long-motion}, by constructing graphs with different marginal distributions specified by different conditions
$\vy[f_j], \vy[i]$, we can generate complex motion sequences.

\subsection{Generation with complex graphs}\label{sec:exp-complex-graph}

\begin{figure}[t!]
    \centering
    \adjincludegraphics[width=\linewidth,trim={{1.9cm} {8.5cm} {1.9cm} {8.5cm}},clip]{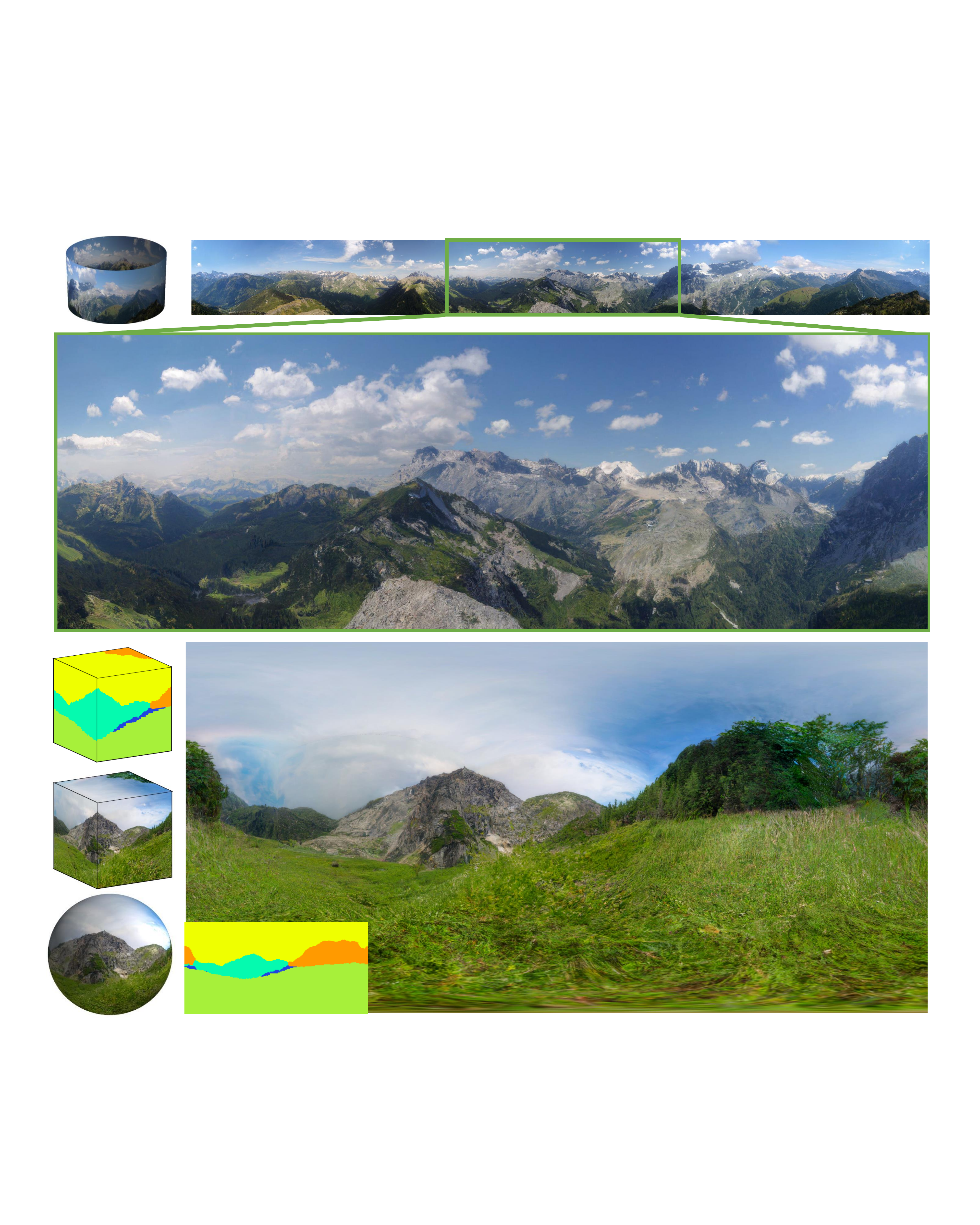}
    \caption{
        Top: a $1024\times 10240$ horizontal panorama image. Bottom left: spherical/cube map representations of an input segmentation map and the output 360-degree panorama image. Each face of the cube is of size $1024\times 1024$. Bottom right: equirectangular representation of the input segmentation and the output image.
    }
    \label{fig:complex-graph}
\end{figure}

We further show that \our{} is able to generate data with a challenging dependency structure specified by a complex graph (such as the ones in \cref{fig:complex-graph}). 
As shown in~\cref{fig:complex-graph}~(top), \our{} can generate a horizontal panorama by constructing a cycle graph.
We also apply our method to generate a 360-degree panorama using a diffusion model trained only on normal perspective images conditioned on semantic segmentation maps~(\cref{fig:complex-graph} bottom). This allows users to create beautiful panoramas from simple doodles, similar to some existing applications such as GauGAN~\cite{park2019semantic} and GauGAN2~\cite{huang2022poegan} but providing a more immersive experience to users. 

\begin{table}[]
\centering
\scalebox{0.82}{
\begin{tabular}{ccccc}
    \toprule
    \multicolumn{1}{c}{Method} & \multicolumn{1}{c}{\begin{tabular}[c]{@{}c@{}}R Precision \\ (top 3)$\uparrow$\end{tabular}} & \multicolumn{1}{c}{FID$\downarrow$} & \multicolumn{1}{c}{\begin{tabular}[c]{@{}c@{}}Multimodal\\ Dist$\downarrow$\end{tabular}} & \multicolumn{1}{c}{Diversity$\rightarrow$} \\
    \midrule
    \cellcolor{lightgray}Real data     & \cellcolor{lightgray}0.798 & \cellcolor{lightgray}0.001 & \cellcolor{lightgray}2.960 & \cellcolor{lightgray}9.471\\
    \cellcolor{lightgray}MDM~\cite{tevet2022human} & \cellcolor{lightgray}0.605 & \cellcolor{lightgray}0.492 & \cellcolor{lightgray}5.607 & \cellcolor{lightgray}9.383\\
    Baseline       & 0.298 & 10.690  & 7.512 & 6.764\\
    Replacement    & 0.567 &  1.281  & 5.751 & 9.184\\
    Reconstruction & 0.585 &  1.012  & 5.716 & 9.175\\
    Ours         & \textbf{0.611} &  \textbf{0.605}  & \textbf{5.569} & \textbf{9.372}\\
    \bottomrule
\end{tabular}
}
\caption{
Quantitative results of long-duration generation on the HumanML3D test set~\cite{Guo_2022_CVPR}.
Dark-colored rows are the results of short-duration motion samples (for reference only), while other rows evaluate methods that generate 24 seconds of motion, which is around 4 times longer than the average length of training data. All methods are based on pre-trained MDM~\cite{tevet2022human}.
$\rightarrow$ means the results are better if the metric is closer to real data.
}
\label{tab:motion-seq-perf}
\end{table}

\section{Conclusion}

In this work, we propose \ours, a novel diffusion model that can synthesize large content via a collection of diffusion models trained on pieces of large content. \ours is based on factor graph representation and inspired by Bethe approximation, both commonly used in probabilistic graphical models.
\ours is scalable; it allows different diffusion models trained only with samples from marginal distributions instead of joint data distribution, which are easier to obtain.
\ours is efficient; diffusion models for different marginals can be trained and sampled in parallel. Through \our{}, we enable large content generation with diffusion models.

\newpage

{\small
\bibliographystyle{ieee_fullname}
\bibliography{egbib,mendeley}
}

\clearpage
\onecolumn
\appendix

\section{Factor graph and Bethe approximation}

\paragraph{Approximations}
The Bethe approximation is a popular technique used in variational inference, probabilistic graphical models, and density estimation. 
One of the key pillars of Bethe approximation is approximating distribution entropy with Bethe entropy~\cite{KolFri09}. 
In terms of the factor graph, 
\begin{equation}~\label{eq:bethe-entropy}
    \gH_B := \sum_{j=1}^m \gH_f^{(j)} + \sum_{i=1}^n (1 - d_i) \gH_\vx^{(i)},
\end{equation}
where $\gH_f^{(j)}, \gH_\vx^{(i)}$ denote entropy of marginal distribution $q(f^{(j)}), q(\vx^{(i)})$ defined over factor nodes and variable nodes respectively, that is,
\[
    \gH_f^{(j)} =\int_{f^{(j)}} -q(f^{(j)}) \log q(f^{(j)}) df^{(j)}, \quad
    \gH_\vx^{(i)} =\int_{\vx^{(i)}} -q(\vx^{(i)}) \log q(\vx^{(i)}) d\vx^{(i)}.
\]
In fact,~\cref{eq:bethe-entropy} can be derived via the joint distribution approximation by its marginal distribution, 
\begin{equation}
    \int_\vu -p(\vu) \log p(\vu) d\vu = 
    \sum_{j=1}^m \int_{f^{(j)}} -q(f^{(j)}) \log q(f^{(j)}) df^{(j)}
    + \sum_{i=1}^n (1-d_i) \int_{\vx^{(i)}} -q(\vx^{(i)}) \log q(\vx^{(i)}) d\vx^{(i)}.
\end{equation}
Approximations~\cref{eq:bethe-entropy} are exact when the factor graph is an acyclic graph while practitioners also use approximations in general graphs where loops appear in the graph.

\paragraph{Expressiveness of \ours}
The expressiveness of $p_{\theta}$ heavily depends on the underlying graphical model, which encodes the conditional independence structure of data in the sparse graph structure.
A sparser graph results in a simpler joint distribution from which it is easier to draw samples. 
The sparsest graph consists of only variable nodes and no factor nodes, which indicates that all random variables are independent. 
Though it is the simplest joint distribution, such models fail to capture correlations between random variables and cannot express common distributions in the real world. 
On the other hand, while the fully connected graphs posses rich representation ability, they are difficult to infer or generate samples from, and Bethe approximation may suffer from large bias.
Empirically, \ours is expressive enough to approximate complex distributions on real datasets with different modalities. 

\paragraph{Hierarchical factor graph}
Though we only present several simple factor graphs in the main paper, we can construct more expressive graphical models with hierarchical factor graphs. The idea behind a ``hierarchical'' factor graph is to treat the nodes and factors themselves as joint distributions over multiple random variables; by modeling these nodes / factors (defined over multiple random variables) with factor graphs, and their likelihood can still be evaluated with Bethe approximation.

The generation of 360-degree images is a good example, where each factor node in 360 cubemap graph%
is a smaller factor graph. Concretely, let us denote 6 faces as $\vx^{\text{(F)}}, \vx^{\text{(B)}}, \vx^{\text{(L)}}, \vx^{\text{(R)}}, \vx^{\text{(U)}}, \vx^{\text{(D)}}$. 
Then the three factor nodes follow $f^{(1)} = \{  
\vx^{\text{(F)}}, \vx^{\text{(B)}}, \vx^{\text{(L)}}, \vx^{\text{(R)}}
\}$,
$f^{(2)} = \{  
\vx^{\text{(F)}}, \vx^{\text{(B)}}, \vx^{\text{(U)}}, \vx^{\text{(D)}}
\}$,
$f^{(3)} = \{  
\vx^{\text{(L)}}, \vx^{\text{(R)}}, \vx^{\text{(U)}}, \vx^{\text{(D)}}
\}$.
The three variable nodes are 
$\vx^{(1)} = \{ \vx^{\text{(L)}}, \vx^{\text{(R)}} \} $,
$\vx^{(2)} = \{ \vx^{\text{(U)}}, \vx^{\text{(D)}} \} $,
$\vx^{(3)} = \{ \vx^{\text{(F)}}, \vx^{\text{(B)}} \}$.
We can apply Bethe approximation for the factor graph over $\vx^{(1)},\vx^{(2)},\vx^{(3)}$ and $f^{(1)}, f^{(2)}, f^{(3)}$. For the individual nodes and factors, the likelihood is defined over a set of random variables that can be modeled with factor graphs again, this time over the 6 faces. For the factor nodes, we treat it as a loop graph; for the variable nodes, as the faces contained within are opposite to each other, the corresponding factor graph would be just the two disconnected components.

Another way to improve expressiveness is to incorporate different conditional signals in different nodes of the factor graph. 
The approach can be interpreted as another type of hierarchical factor graph, which involves latent codes based on different conditional signals. In fact, several works~\cite{lin2019coco,skorokhodov2021aligning} that generate different patches of large images independently based on global code can be viewed as a hierarchical factor graph with latent code. We apply similar techniques in our conditional generation tasks, such as text-conditioned motion generation, and segmentation-conditioned image generation.

For applications that involve more complex dependencies among random variables and demand difficult inference tasks, more general graph representation, such as Junction tree~\cite{KolFri09}, may have some advantages over representations based on factor graphs. 
We leave the generalization of \ours to more complicated graphs for future research.

\section{Training and Sampling of \ours}

\subsection{Training}
\ours demands diffusion models over different pieces of the target content. 
Ideally, \ours can work out of the box if diffusion models over each node in the factor graphs are available.
When pre-trained models are not accessible, we can train \ours in the same way as training standard diffusion models. 
We list the training algorithm in~\cref{algo:dc-training,algo:diffusion-node}.
We note that the learning process of one marginal is independent of others, making the training procedure easy to scale since different marginals of \ours can be learned in parallel. Moreover, different variable nodes or factor nodes may share the same diffusion models due to symmetry, improving the scalability further.

\begin{algorithm}[H]
    \caption{Diffuison Collage: Training }
    \begin{algorithmic}
    \label{algo:dc-training}
    \STATE \textbf{Inputs}: Marginal data on factor node $\{\gD[f^{(j)}]\}$, marginal data on variable node $\{\gD[i]\}$
    \STATE \textbf{Output}: Score models $\vs_\theta$ for marginal distributions
    \STATE \codecomment{\# Training for marginals can be conducted in parallel.}
    \FOR{$j \in {1,2,\cdots, m}$}  
        \STATE Training diffusion model $\vs_\theta(f^{(j)}, t)$ on data $\gD[f^{(j)}]$
    \ENDFOR
    \FOR{$i \in {1,2,\cdots, n}$}  
        \STATE Training diffusion model $\vs_\theta(\vx^{(i)}, t)$ on data $\gD[i]$
    \ENDFOR
    \end{algorithmic}
\end{algorithm}

\begin{algorithm}[H]
    \caption{Training diffusion models for one node}
    \begin{algorithmic}
    \label{algo:diffusion-node}
    \STATE \textbf{Inputs}: Marginal data $\gD$
    \STATE \textbf{Output}: Score models $\vs_\theta$
    \STATE \textbf{repeat}
        \STATE\hspace{\algorithmicindent} Sample $\vu_0$ from $\gD$ 
        \STATE\hspace{\algorithmicindent} Sample $t$ and Gaussian noise $\epsilon$
        \STATE\hspace{\algorithmicindent} $\vu_t = \vu_0 + \sigma_t \epsilon$ 
        \STATE\hspace{\algorithmicindent} Gradient descent on 
            $\nabla_\theta  [\omega(t) \norm{
                 \nabla_{\vu_t} \log q_{0t}(\vu_t | \vu_0) -
                 \vs_\theta(\vu_t, t)
            }^2]$
    \STATE \textbf{util} converged
    \end{algorithmic}
\end{algorithm}

\subsection{Sampling}

After training diffusion models for each marginal, \ours implicitly obtains $p_\theta(\vu, t)$ by its score $\nabla \log p_\theta(\vu,t)$.
The score of the learned distribution can be composited with its marginal scores $\vs_\theta(\vx^{(i)}, t),\vs_\theta(f^{(j)}, t)$:
\begin{align}\label{eq:dc-app-score}
    \nabla \log p_\theta(\vu,t) = \vs_\theta(\vu,t) = \sum_{j=1}^m \vs_\theta(f^{(j)},t) + \sum_{i=1}^n (1 - d_i) \vs_\theta(\vx^{(i)},t).
\end{align}
The marginal scores can be computed in parallel over the entire large content, which would significantly reduce the latency of the algorithm.
We can plug~\cref{eq:dc-app-score} into existing diffusion model sampling algorithms. We include a deterministic sampling algorithm in~\cref{algo:dc-sampling} for reference, though we re-emphasize that any sampler applicable to regular diffusion models would work with \ours{}. Besides, \ours also inherits the versatility of diffusion models and allows controllable generation without re-training, such as inpainting and super-resolution~\cite{choi2021ilvr,kawar2022denoising}.
We include more details regarding training-free conditional generation in~\cref{sec:rep-recon}.

\begin{algorithm}[H]
    \caption{\ours{}: Sampling with Euler}
    \begin{algorithmic}
    \label{algo:dc-sampling}
    \STATE \textbf{Inputs}: Score models $\vs_\theta$, decreasing time steps $\{ t_k\}_{k=0}^K$
    \STATE \textbf{Output}: Samples from $p_\theta(\vu)$
    \STATE Sample $\vu_K$ from prior distribution $\gN(0, \sigma_{t_K} \mI)$
    \FOR{$k \in {K, K-1, \cdots 1}$}
        \STATE \codecomment{\# Pieces of $\vs_\theta(\vu_k, t_k)$ can be evaluated in parallel.}
        \STATE $\vu_{k-1} = \vu_k + \dot{\sigma}_{t_k} {\sigma}_{t_k} \vs_\theta(\vu_k,t_k)(t_k - t_{k-1})$
    \ENDFOR
    \STATE Return $\vu_0$
    \end{algorithmic}
\end{algorithm}

\section{Experiments details}

\subsection{Replacement and Reconstruction Methods for Conditioning}\label{sec:rep-recon}
Here, we describe the details of \textbf{replacement} and \textbf{reconstruction} methods that are compared with \ours in the experiments. In both cases, we are provided with an extra condition $\vy$, and our goal is to generate $\vu$ such that $\vy = H(\vu)$ for some known function $H$. For example, $H: \mathbb{R}^{n} \to \mathbb{R}^{m}$ can be a low-pass filter that produces a low-resolution image (dimension $m$) from a high-resolution image (dimension $n$), and the task would essentially become super-resolution; similarly, one could define an inpainting task where $H$ is taking a subset of the pixels of the image $\vx$. Diffusion models are particularly better-suited to such inverse problems than other generative models, such as GANs~\cite{pan2021exploiting}, as they can produce good results with much fewer iterations~\cite{kawar2022denoising}. 

Both replacement and reconstruction methods make some modifications to the sampling procedure. At a high level, the replacement method makes a prediction over the clean image (denoted as $\hat{\vu}_0$), and replaces parts of the image $\hat{\rvu}_0$ using information about $\vy$; one could implement this as a projection if $H \in \mathbb{R}^{m \times n}$ is a matrix, \textit{i.e.}, $\mathrm{proj}(\hat{\vu}_0) = H^\dagger \vy + (I - H^\dagger H) \hat{\vu}_0$ where $H^\dagger$ is the pseudoinverse of $H$. This is the strategy used in ILVR~\cite{choi2021ilvr} and DDRM~\cite{kawar2022denoising}. The reconstruction method, on the other hand, takes an additional gradient step on top of the existing sampling step that minimizes the $L_2$ distance between $\vy$ and $H \hat{\rvu}_0$; this has been shown to produce higher-quality images than replacement methods on super-resolution and inpainting~\cite{chung2022improving}. 
We describe the two types of conditional sampling algorithms in \cref{algo:replacement-sampling} and \cref{algo:reconstruction-sampling}, respectively, using \ours{}. 
This is almost identical to the conditional sampling algorithms with a standard diffusion model, as we only changed the diffusion to the one constructed by \ours{}. 
For autoregressive baselines, %
we use these algorithms with regular diffusion models; for inpainting experiments with large images, %
we use them with \ours{}.

\begin{algorithm}[H]
    \caption{Replacement-based Conditioning using Regular Diffusion Models}
    \begin{algorithmic}
    \label{algo:replacement-sampling}
    \STATE \textbf{Inputs}: Observation $\vy$, matrix $H$, score models $\vs_\theta$, decreasing time steps $\{ t_k\}_{k=0}^K$, sampling algorithm from time $t$ to time $s$ using a score function, denoted as $\text{sample}(\text{score}, \vu_t, t, s)$.
    \STATE \textbf{Output}: Samples from $p_\theta(\vu)$ where $\vy = H(\vu)$
    \STATE Sample $\vu_K$ from prior distribution $\gN(0, \sigma_{t_K} \mI)$
    \FOR{$k \in {K, K-1, \cdots 1}$}
        \STATE \codecomment{\# Obtain denoising result from score function $\vs_\theta(\vu_k, t_k)$.}
        \STATE $\hat{\vu}_0 = \vu_k + \sigma_{t_k}^2 \vs_\theta(\vu_k, t_k)$.
        \STATE \codecomment{\# Replacement projection based in $\vy$ and $H$.}
        \STATE $\tilde{\vu}_0 = H^\dagger \vy + (I - H^\dagger H) \hat{\vu}_0$.
        \STATE \codecomment{\# Sample based on corrected result.}
        \STATE $\tilde{\vs} = (\tilde{\vu}_0 - \vu_{k}) / \sigma_{t_k}^2$.
        \STATE $\vu_{k-1} = \text{sample}(\tilde{\vs}, \vu_{t_k}, t_k, t_{k-1})$.
    \ENDFOR
    \STATE Return $\vu_0$
    \end{algorithmic}
\end{algorithm}

\begin{algorithm}[H]
    \caption{Reconstruction-based Conditioning with \ours{} or Regular Diffusion Models}
    \begin{algorithmic}
    \label{algo:reconstruction-sampling}
    \STATE \textbf{Inputs}: Observation $\vy$, matrix $H$, score models $\vs_\theta$, decreasing time steps $\{ t_k\}_{k=0}^K$, sampling algorithm from time $t$ to time $s$ using a score function, denoted as $\text{sample}(\text{score}, \vu_t, t, s)$, and hyperparameter for reconstruction gradient $\lambda_t$.
    \STATE \textbf{Output}: Samples from $p_\theta(\vu)$ where $\vy = H(\vu)$
    \STATE Sample $\vu_K$ from prior distribution $\gN(0, \sigma_{t_K} \mI)$
    \FOR{$k \in {K, K-1, \cdots 1}$}
        \STATE \codecomment{\# Obtain denoising result from score function $\vs_\theta(\vu_k, t_k)$.}
        \STATE $\hat{\vu}_0 = \vu_k + \sigma_{t_k}^2 \vs_\theta(\vu_k, t_k)$.
        \STATE \codecomment{\# Update score based in $\vy$ and $H$.}
        \STATE $\tilde{\vs} = \vs_\theta(\vu_k, t_k) + \lambda_t \nabla_{\vu_k} \Vert H \hat{\vu}_0 - \vy\Vert_2^2$.
        \STATE \codecomment{\# Sample based on new score function.}
        \STATE $\vu_{k-1} = \text{sample}(\tilde{\vs}, \vu_{t_k}, t_k, t_{k-1})$.
    \ENDFOR
    \STATE Return $\vu_0$
    \end{algorithmic}
\end{algorithm}

\subsection{Image experiments}

\begin{algorithm}[H]
    \caption{Inifnite image generation with \ours{}: training}
    \begin{algorithmic}
    \label{algo:dc-inf-training}
    \STATE \textbf{Inputs}: Square image data $\gD$
    \STATE \textbf{Output}: Shift-invariant score model $\vs_\theta$ for both factor nodes and variable nodes
    \STATE \textbf{repeat}
        \STATE\hspace{\algorithmicindent} Sample $\vu_0$ from $\gD$
        \STATE\hspace{\algorithmicindent} Random crop $\vu_0$ by half with $50 \%$ probability
        \STATE\hspace{\algorithmicindent} Sample $t$ and Gaussian noise $\epsilon$ with shape of $\vu_0$
        \STATE\hspace{\algorithmicindent} $\vu_t = \vu_0 + \sigma_t \epsilon$ 
        \STATE\hspace{\algorithmicindent} Gradient descent on 
            $\nabla_\theta [\omega(t) \norm{
                 \nabla_{\vu_t} \log q_{0t}(\vu_t | \vu_0) -
                 \vs_\theta(\vu_t, t)
            }^2]$
    \STATE \textbf{util} converged
    \end{algorithmic}
\end{algorithm}

\paragraph{Training}
To finetune GLIDE~\cite{Nichol2021a} on our internal dataset, we first train our base $64\times 64$ model with a learning rate $1\times 10^{-4}$ and a batch size 128 for $300K$ iterations.
Then we finetune $64 \rightarrow 256, 256\rightarrow 1024$ upsamplers for $100K, 50K$ iterations.
For the $256\rightarrow 1024$ upsampler, we finetune the upsampler of eDiff-I~\cite{balaji2022ediffi}. 
Following the prior works~\cite{balaji2022ediffi,saharia2022photorealistic}, we train the $256\rightarrow 1024$ model using random patches of size $256{\times}256$ during training and apply it on $1024{\times}1024$ resolution during inference. 
We utilize AdamW optimzer~\cite{loshchilov2017decoupled} and apply exponential moving average~(EMA) with a rate $0.999$ during training. 
The base $64 \times 64$ diffusion model is trained to be conditioned on image CLIP embeddings with a random drop rate $50 \%$ while the two upsampling diffusion models are only conditioned on low-resolution images.
For the diffusion model conditioned on semantic segmentation maps, we replace the first layer of our pre-trained base $64\times 64$ model and concatenate embeddings of semantic segmentation maps and noised image inputs. 
We further finetune the diffusion model for another $100K$ iterations conditioned on segmentation.

For experiments on LHQ~\cite{skorokhodov2021aligning} and LSUN~\cite{yu2015lsun} Tower, we train diffusion models from scratch with the U-net architecture proposed in Dhariwal~\etal~\cite{dhariwal2021diffusion}. 
Thanks to its success in LSUN and ImageNet~\cite{deng2009imagenet}, we adopt its hyperparameters for LSUN dataset in \cite[Table 11]{dhariwal2021diffusion}.
Due to limited computational resources, we train diffusion models with channel size 192 and batch size 128 for $100K$ iterations instead of the recommended hyperparameters. We follow the data preprocessing in Skorokhodov~\etal~\cite[Algorithm 1]{skorokhodov2021aligning} with its official implementation~\footnote{\url{https://gist.github.com/universome/3140f74058a48aa56a556b0d9e24e857}}, which extracts a subset with approximately horizontally invariant statistics from original datasets. 

Thanks to the shift-invariant property of infinite images, we use the same diffusion model to fit both factor and variable nodes, where the width of images over the variable node is half of the width of factor nodes. The dataset for variable nodes consists of random cropped images from factor nodes. We list its training algorithm in~\cref{algo:dc-inf-training}. We apply a similar strategy to train segmentation-conditioned diffusion models. 
We adopt VESDE and preconditioners proposed in Karras \etal~\cite{karras2022elucidating} to train our diffusion models.

\paragraph{Sampling}
Regarding sampling image diffusion models, we use 
the stochastic sampler in Karras \etal~\cite{karras2022elucidating}
with $80$ sampling steps and default hyperparameters. 
We find stochastic samplers are slightly better than deterministic samplers in \ours.
For quantitative comparison on our internal dataset, we have the same CLIP embedding for both factor and variable nodes in one graph while we use unconditional generation on LHQ and LSUN Tower.
We use the same sampler for baseline and autoregressive methods based on replacement or reconstruction.
To connect different styles and real images with a linear chain graph, we interpolate conditional signals with spherical linear interpolation~\cite{white2016sampling}. 
We find \ours with~\cref{algo:replacement-sampling} can produce satisfying samples for conditional generation efficiently. 
More visual examples are included in~\cref{sec:additional-sample}.

\subsection{Motion experiments}

We use the pre-trained diffusion model\footnote{\url{https://github.com/GuyTevet/motion-diffusion-model}} from~\cite{tevet2022human} and only make the following modifications during sampling.

\begin{itemize}
\item Similar to experiments in images, we inpaint motion sequences by masking $50\%$ content in the sliding window for Replacement and Reconstruction methods.
\item All experiments employ the deterministic DDIM sampler~\cite{song2020denoising} with 50 steps.
\item We use the same prompt to denoise both factor and variables nodes for long motion experiments benchmark experiments results and~\cref{tab:motion-seq-perf-std}.
\item To composite motions with multiple actions, we decompose the given long prompts into several short sentences manually so that each sentence only consists of one or two actions similar to prompts in the training data. Then we assign each factor $\vy[f_j]$ with one short prompt sequentially and unconditional null token for the variables node.
\item Analogous to circle image generation, we add a factor node connected to the head and tail variable nodes in the factor graph. 
\end{itemize}

We include standard derivation for long motion experiments in~\cref{tab:motion-seq-perf-std}.
\begin{table}[]
\centering
\scalebox{0.82}{
\begin{tabular}{ccccc}
    \toprule
    \multicolumn{1}{c}{Method} & \multicolumn{1}{c}{\begin{tabular}[c]{@{}c@{}}R Precision \\ (top 3)$\uparrow$\end{tabular}} & \multicolumn{1}{c}{FID$\downarrow$} & \multicolumn{1}{c}{\begin{tabular}[c]{@{}c@{}}Multimodal\\ Dist$\downarrow$\end{tabular}} & \multicolumn{1}{c}{Diversity$\rightarrow$} \\
    \midrule
    Real data   & 0.798$\pm$0.002 & 0.001$\pm$0.000 & 2.960$\pm$0.006 &9.471$\pm$0.100\\
    MDM~\cite{tevet2022human} & 0.605$\pm$0.005 & 0.492$\pm$0.036 & 5.607$\pm$0.028 &9.383$\pm$0.070\\
    Baseline       & 0.298$\pm$ 0.006 & 10.690 $\pm$0.179 & 7.512$\pm$0.039 & 6.764$\pm$0.069\\
    Replacement    & 0.567$\pm$ 0.008 &  1.281 $\pm$0.177 & 5.751$\pm$0.034 & 9.184$\pm$0.122\\
    Reconstruction & 0.585$\pm$ 0.007 &  1.012 $\pm$0.080 & 5.716$\pm$0.033 & 9.175$\pm$0.120\\
    \our{}         & 0.611$\pm$ 0.004 &  0.605 $\pm$0.082 & 5.569$\pm$0.017 & 9.372$\pm$0.109\\
    \bottomrule
\end{tabular}
}
\caption{
Performance on every metric is reported based on a mean and standard derivation of 20 independent evaluations.
}
\label{tab:motion-seq-perf-std}
\end{table}

\section{Limitations}

Despite the clear advantages that \ours{} has over traditional methods, \ours{} is no silver bullet for every large content generation problem. We discuss some limitations below. %

\paragraph{Conditional independence assumptions.} Since we use diffusion models trained on smaller pieces of the content, \ours{} place conditional independence assumptions over the joint distribution of the large content, similar to autoregressive outpainting methods. Sometimes this assumption is reasonable (such as long images for landscape or ``corgis having dinner at a long table''), but there are cases where the long-range dependency is necessary for generating the content. For example, generating a long image of a snake would be difficult with \ours{}, since we drop the conditional dependencies between the head and the tail of the snake, and it is possible that our snake would have two heads or two tails. Part of this can be mitigated by providing global conditioning information, such as the segmentation maps in landscapes. 

\paragraph{Memory footprint.} We reduce the latency of the long content generation by running the diffusion model computations in parallel, and it comes at a cost of using more peak memory than autoregressive methods. 

\paragraph{Number of steps in the sampler.} To ensure global consistency, information needs to flow through the factor graph. This is done by the sum over the overlapping regions in each iteration, so it can be treated as some kind of ``message passing'' behavior. Similar to ``message passing'', many iterations may be needed if the graph diameter is large (even when some global conditioning information is given). For example, for a linear chain with length $L$, we may need the sampler to run $O(L)$ times to get optimal results. Empirically we also find sampling with our method using very few steps in generating infinite images, such as $35$, may result in artifacts. However, we note that this is still much better than the autoregressive counterpart; for a \ours{} implementation that requires $O(L)$ steps of iteration, the reconstruction/replacement methods would require $O(L \times K)$ steps, where $K$ is the number of iterations for the small diffusion models.

\section{Additional samples}~\label{sec:additional-sample}

We include more high-quality samples and motion videos in our supplementary materials. 

\begin{figure*}
    \centering
    \includegraphics[width=0.86\linewidth]{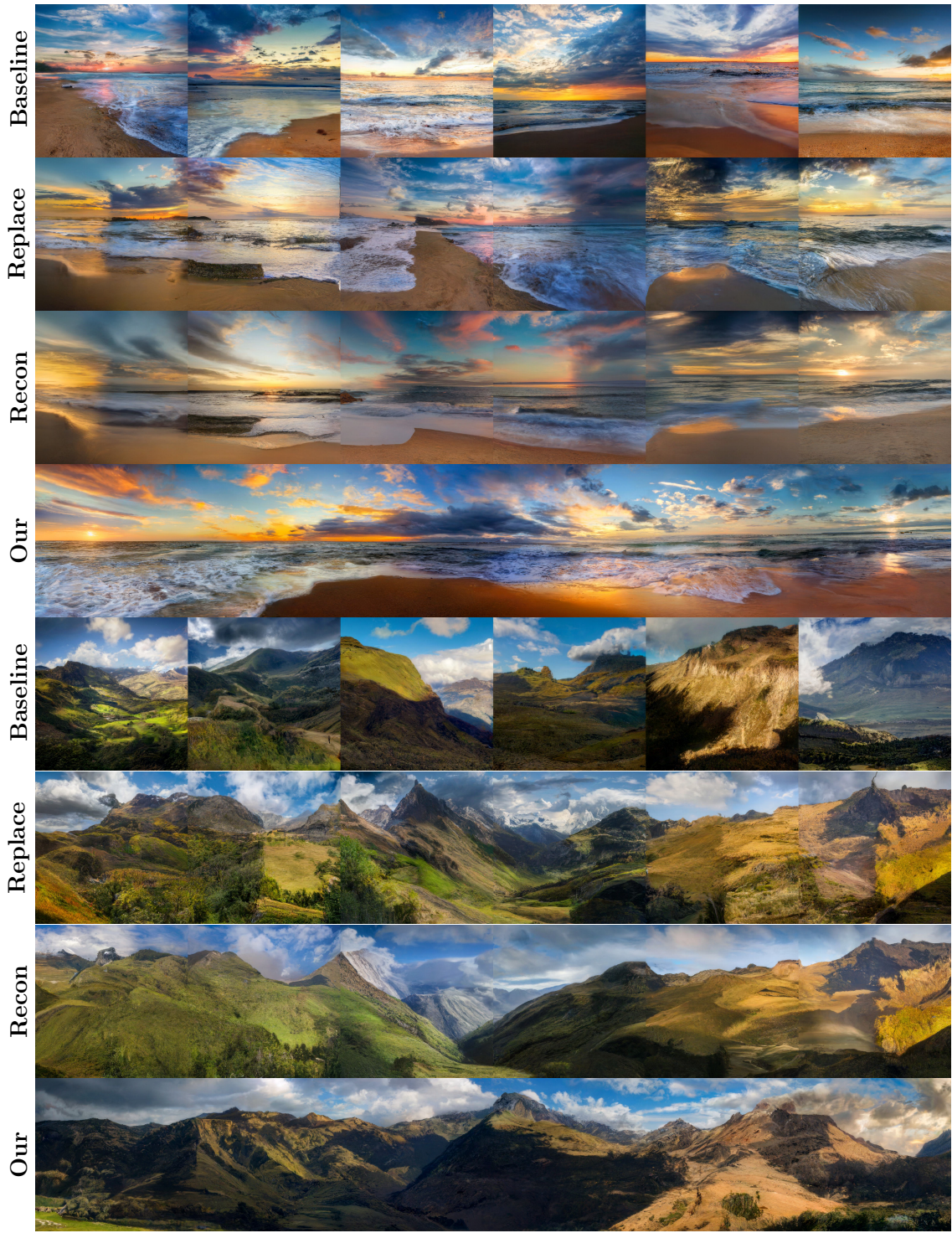}
    \caption{
        More comparison.
    }
\end{figure*}

\begin{figure*}[t!]
    \centering
    \setlength\tabcolsep{0.01cm}
    \begin{tabular}{lcc}

    \raisebox{1.4cm}[0pt][0pt]{\rotatebox[origin=c]{90}{Source}}&\includegraphics[width=0.33\linewidth]{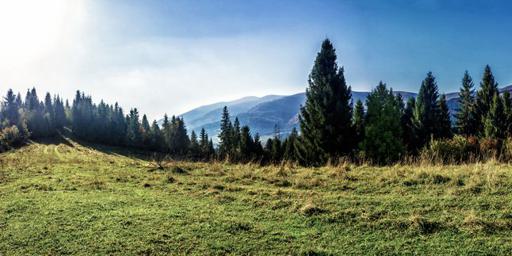}
    &\includegraphics[width=0.66\linewidth]{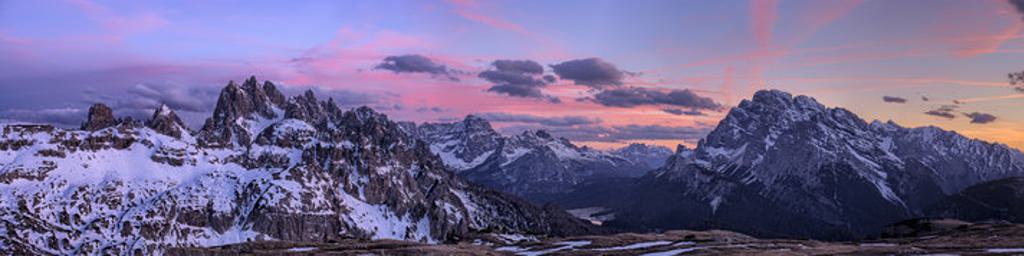} \\

    \raisebox{1.4cm}[0pt][0pt]{\rotatebox[origin=c]{90}{Masked}} \hspace{1em}&\includegraphics[width=0.33\linewidth]{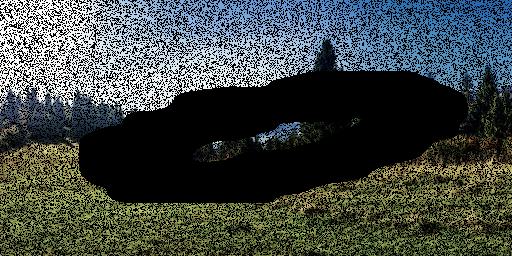}
    &\includegraphics[width=0.66\linewidth]{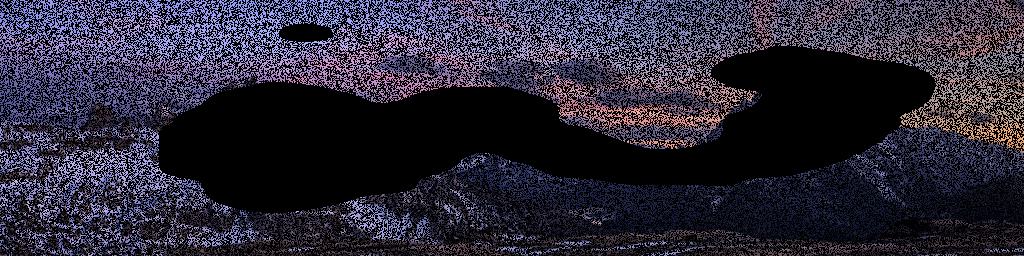} \\

    \raisebox{1.4cm}[0pt][0pt]{\rotatebox[origin=c]{90}{Recon}}&\includegraphics[width=0.33\linewidth]{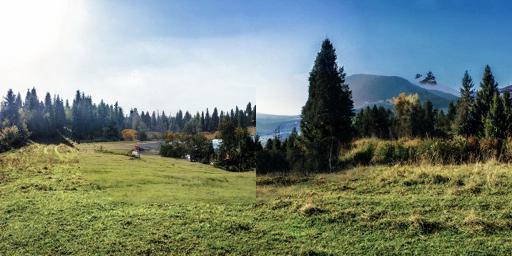}
    &\includegraphics[width=0.66\linewidth]{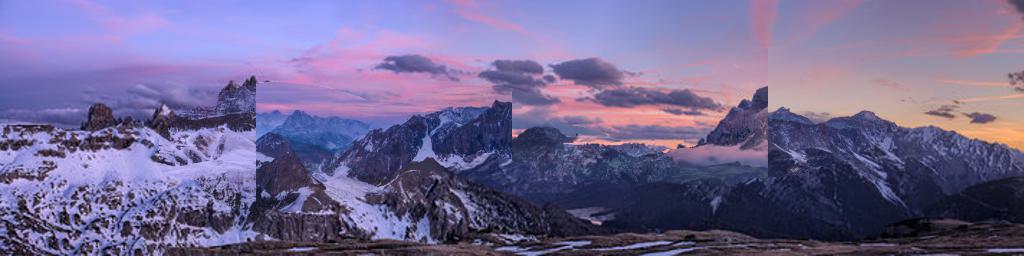} \\

    \raisebox{1.4cm}[0pt][0pt]{\rotatebox[origin=c]{90}{Ours}}&\includegraphics[width=0.33\linewidth]{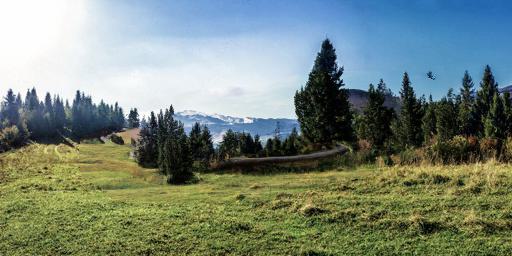}
    &\includegraphics[width=0.66\linewidth]{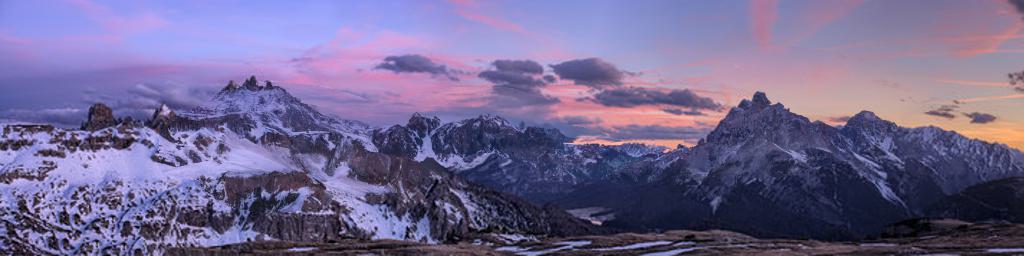} \\
    \end{tabular}
    \caption{
        Inpainting on non-square images. The diffusion models based on smaller patches are run in parallel.
    }
    \label{fig:app-inpainting1}
\end{figure*}

\begin{figure*}[t!]
    \centering
    \setlength\tabcolsep{0.01cm}
    \begin{tabular}{lcc}

    \raisebox{1.4cm}[0pt][0pt]{\rotatebox[origin=c]{90}{Source}}&\includegraphics[width=0.50\linewidth]{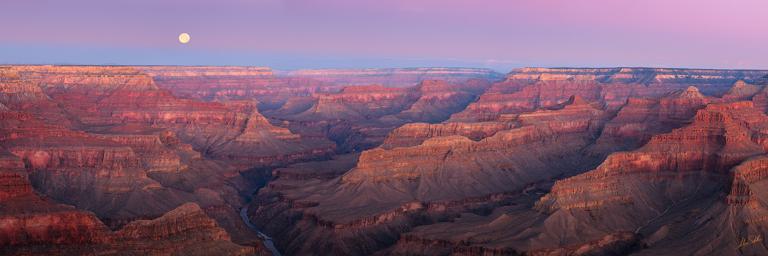}
    &\includegraphics[width=0.50\linewidth]{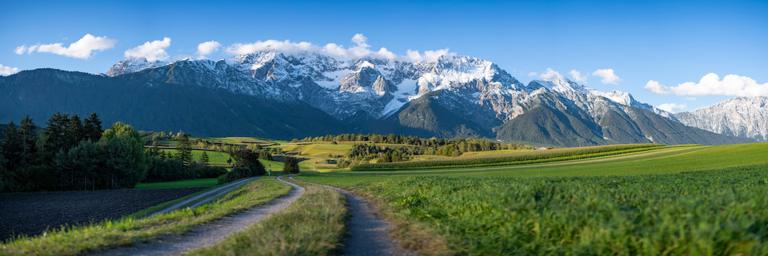} \\

    \raisebox{1.4cm}[0pt][0pt]{\rotatebox[origin=c]{90}{Masked}} \hspace{1em}&\includegraphics[width=0.50\linewidth]{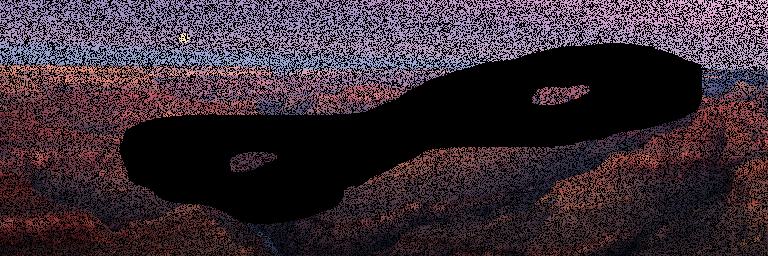}
    &\includegraphics[width=0.50\linewidth]{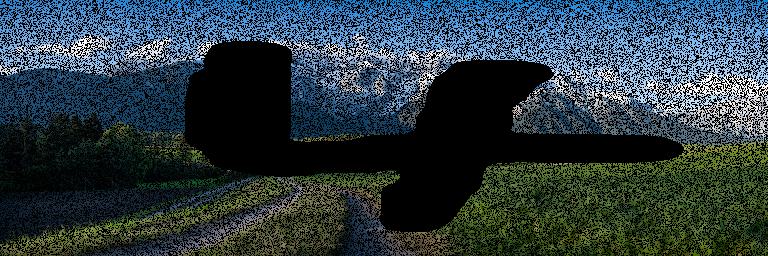} \\

    \raisebox{1.4cm}[0pt][0pt]{\rotatebox[origin=c]{90}{Recon}}&\includegraphics[width=0.50\linewidth]{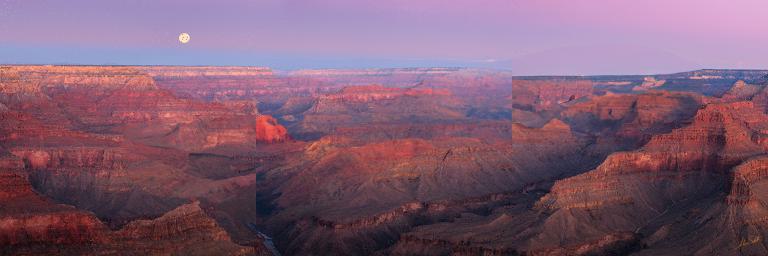}
    &\includegraphics[width=0.50\linewidth]{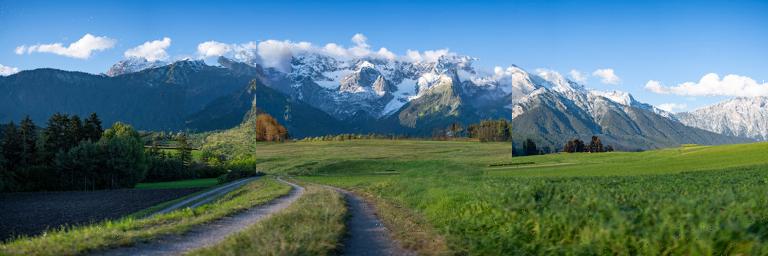} \\

    \raisebox{1.4cm}[0pt][0pt]{\rotatebox[origin=c]{90}{Ours}}&\includegraphics[width=0.50\linewidth]{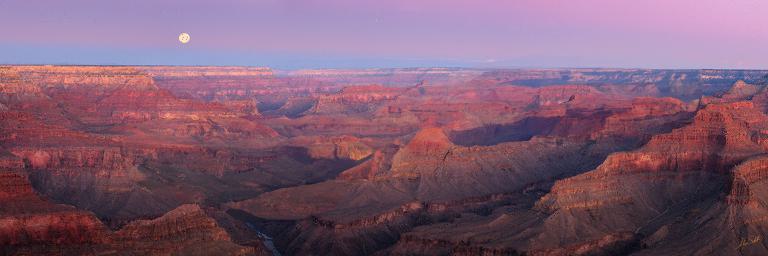}
    &\includegraphics[width=0.50\linewidth]{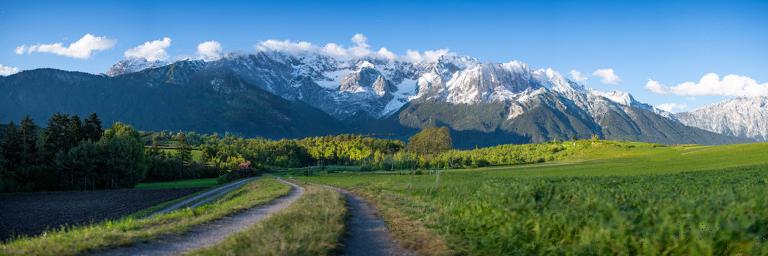} \\
    \end{tabular}
    \caption{
        Inpainting on non-square images. The diffusion models based on smaller patches are run in parallel.
    }
    \label{fig:app-inpainting2}
\end{figure*}

\begin{figure*}
    \centering
    \includegraphics[width=0.82\linewidth]{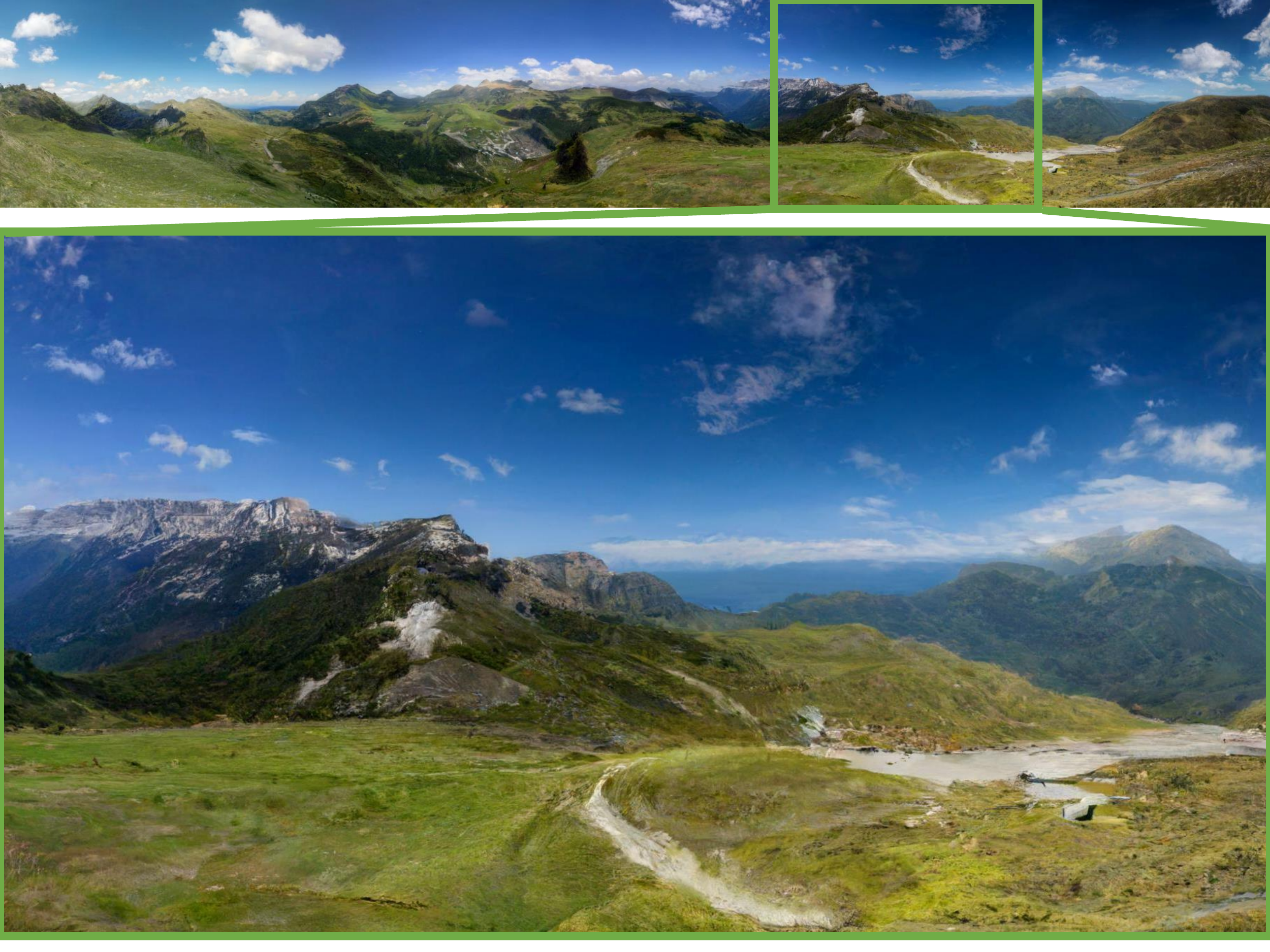}
    \includegraphics[width=0.82\linewidth]{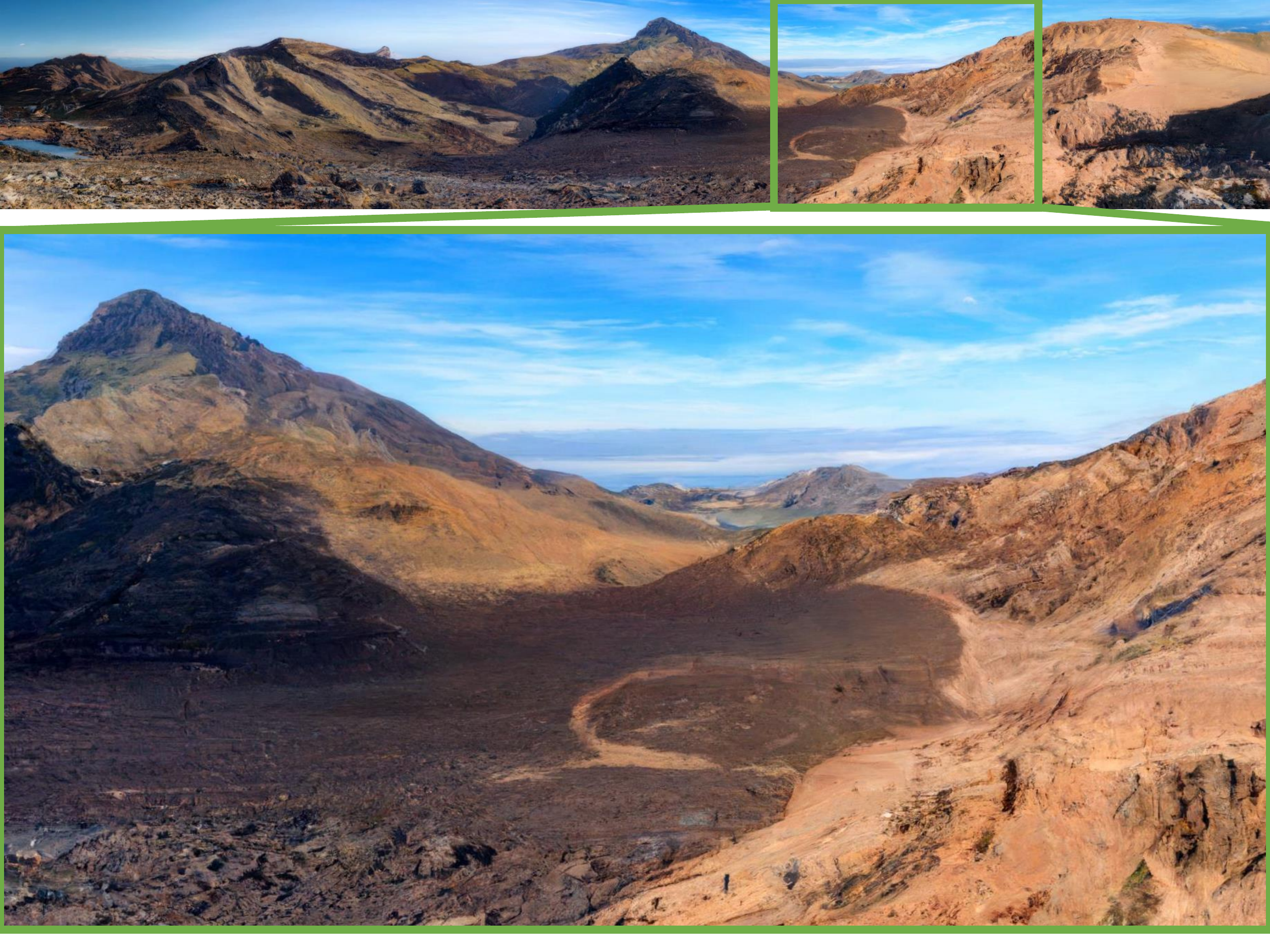}
    \caption{
        \ours on generating long landscape images. Parts are being zoomed in for high-resolution details. 
    }
\end{figure*}

\begin{figure*}
    \centering
    \adjincludegraphics[width=\linewidth,trim={{0cm} {10.5cm} {0cm} {0cm}},clip]{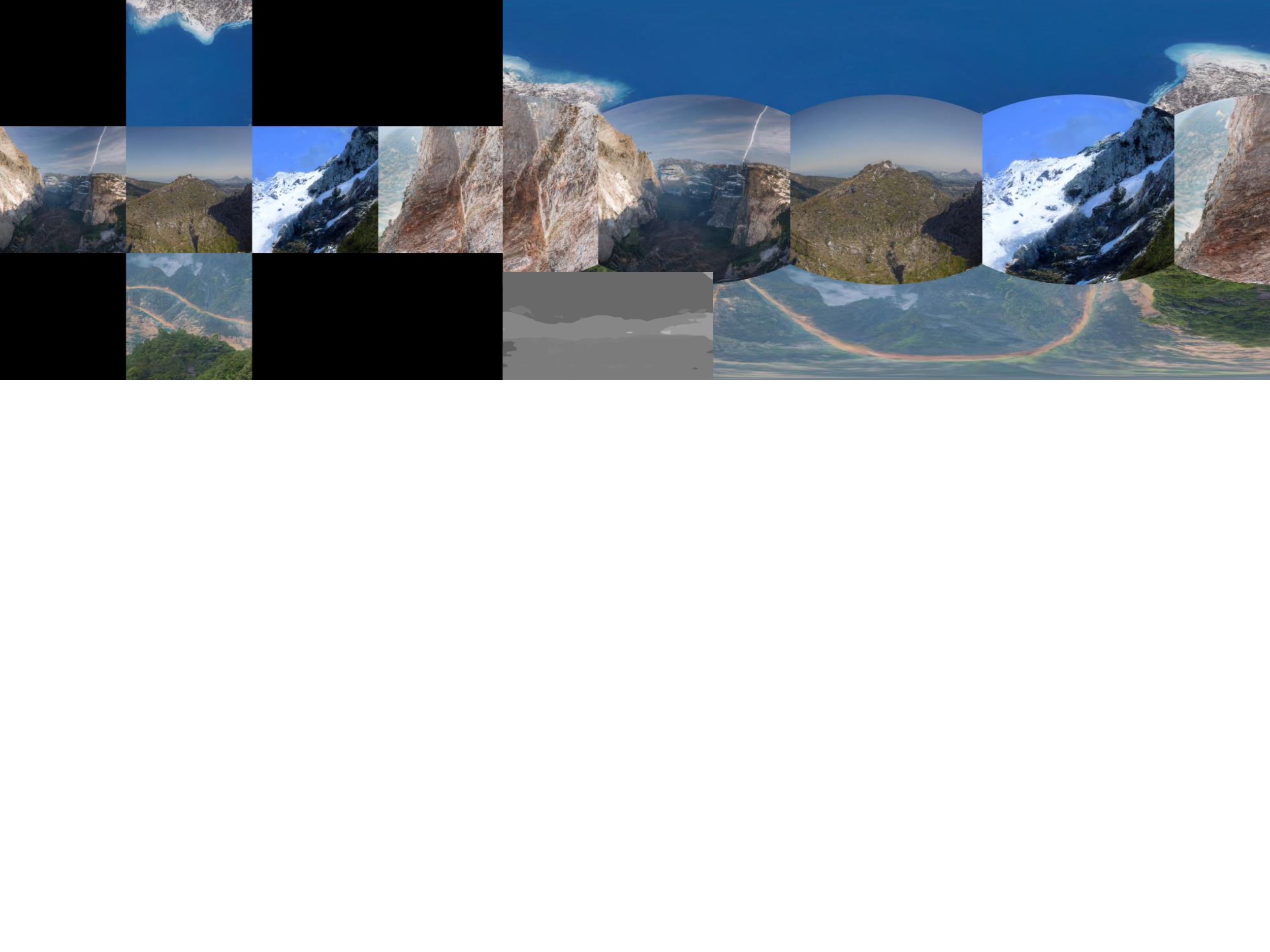}
    \adjincludegraphics[width=\linewidth,trim={{0cm} {10.5cm} {0cm} {0cm}},clip]{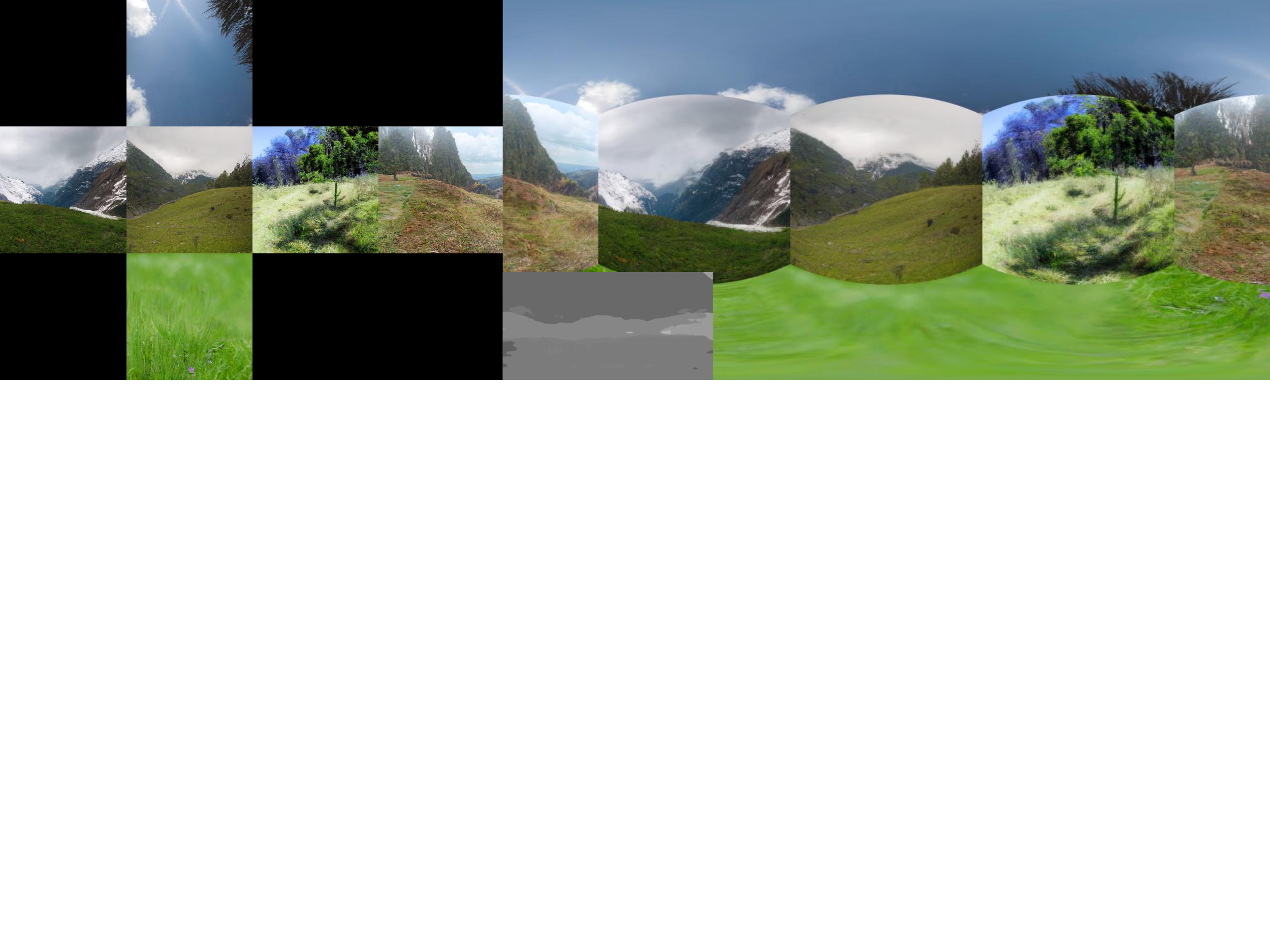}
    \caption{
        Baseline method on 360 cubemap image. (Left) Cubemap representation; (Right) Equirectangular representation; (Lower middle) Semantic segmentation map used to condition the model. Even with a globally-consistent semantic segmentation map, individually processing the patches will lead to the cube faces being quite inconsistent with one another. 
    }
\end{figure*}

\begin{figure*}
    \centering
    \adjincludegraphics[width=\linewidth,trim={{0cm} {10.5cm} {0cm} {0cm}},clip]{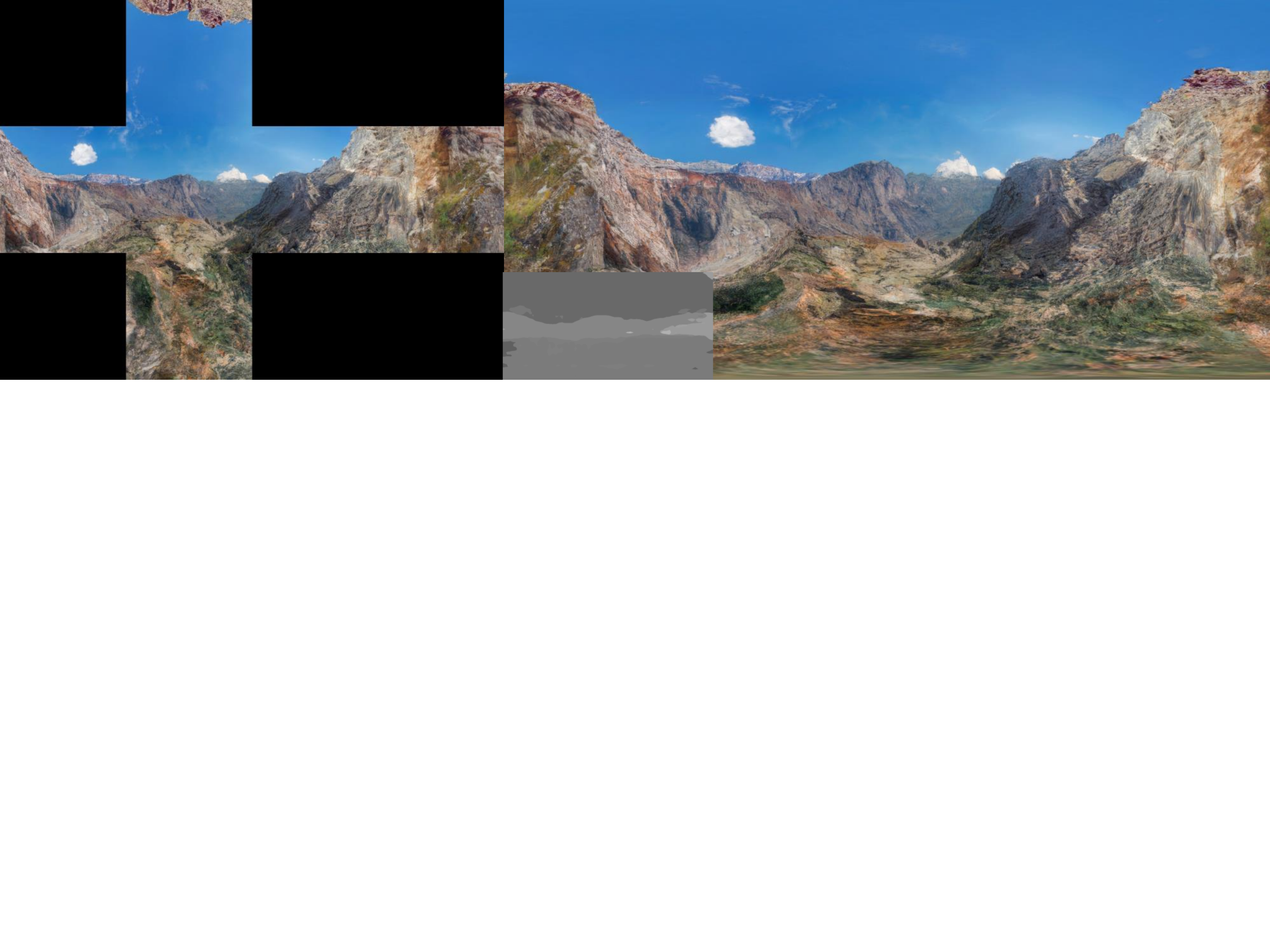}
    \adjincludegraphics[width=\linewidth,trim={{0cm} {10.5cm} {0cm} {0cm}},clip]{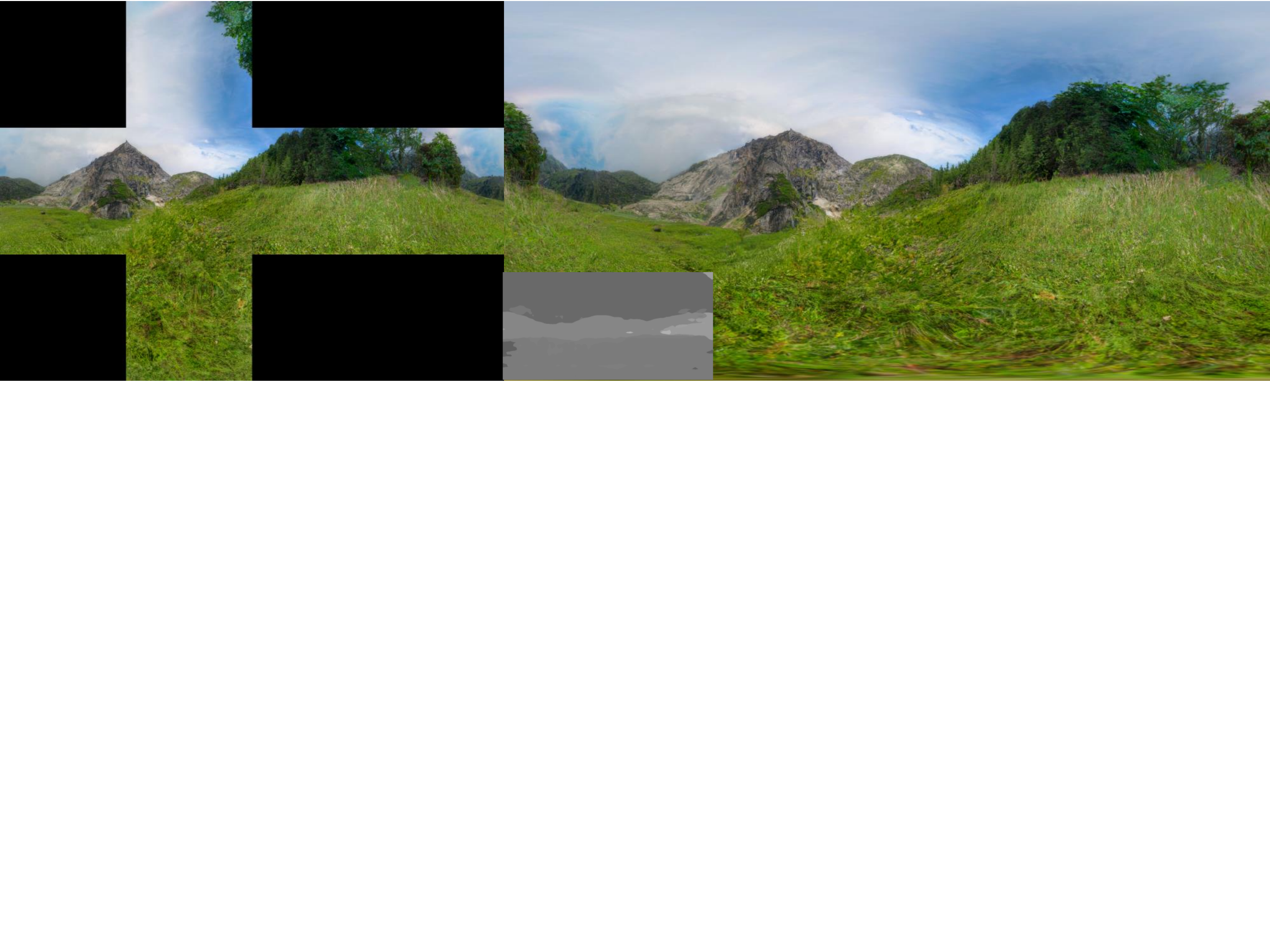}
    \caption{
        \ours on 360 cubemap image. (Left) Cubemap representation; (Right) Equirectangular representation; (Lower middle) Semantic segmentation map used to condition the model. \ours is able to ``connect'' the different faces and produce a globally consistent 360 degree image. 
    }
\end{figure*}

\begin{figure*}
    \centering
    \adjincludegraphics[width=\linewidth,trim={{0cm} {10.5cm} {0cm} {0cm}},clip]{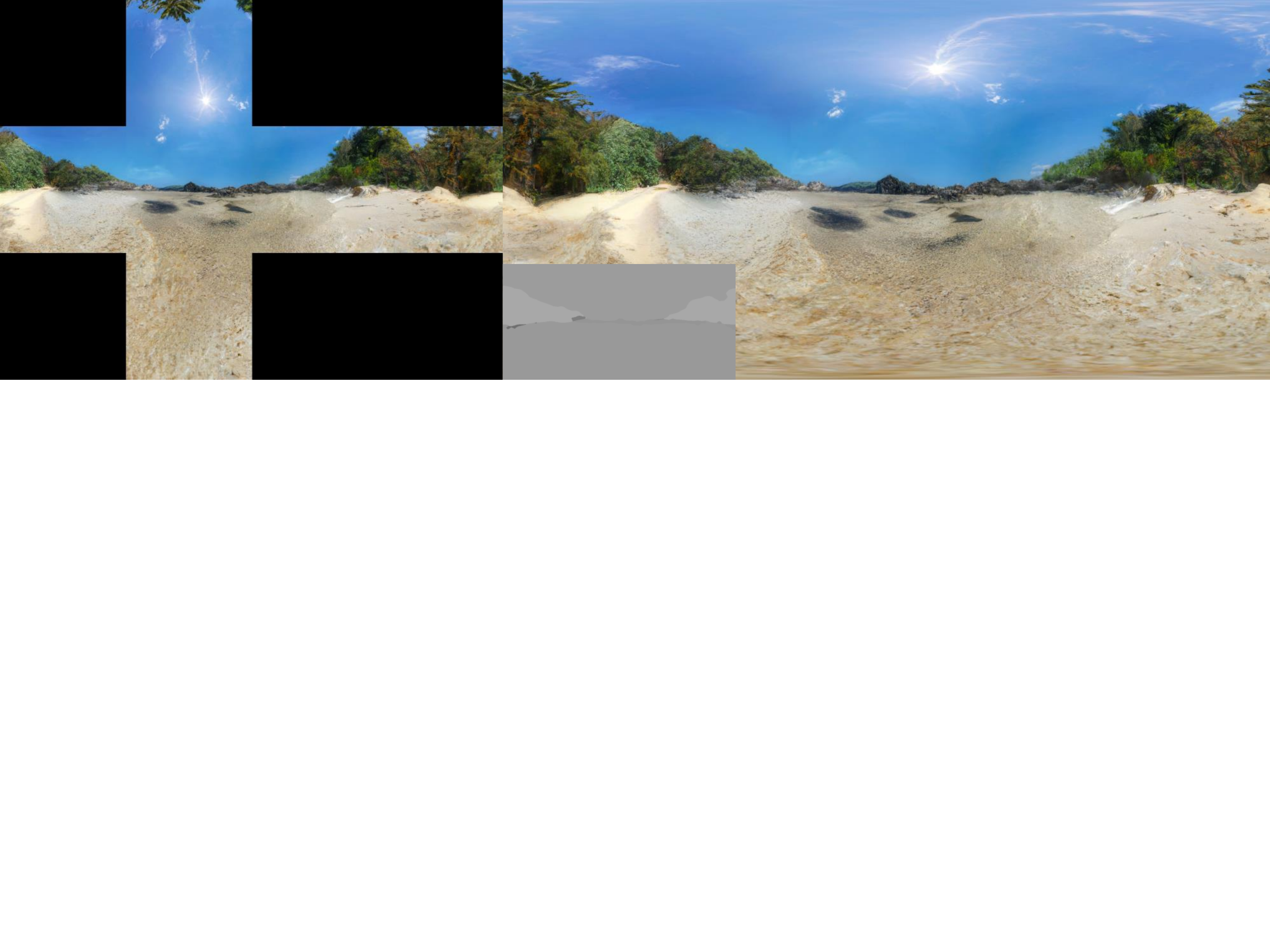}
    \adjincludegraphics[width=\linewidth,trim={{0cm} {10.5cm} {0cm} {0cm}},clip]{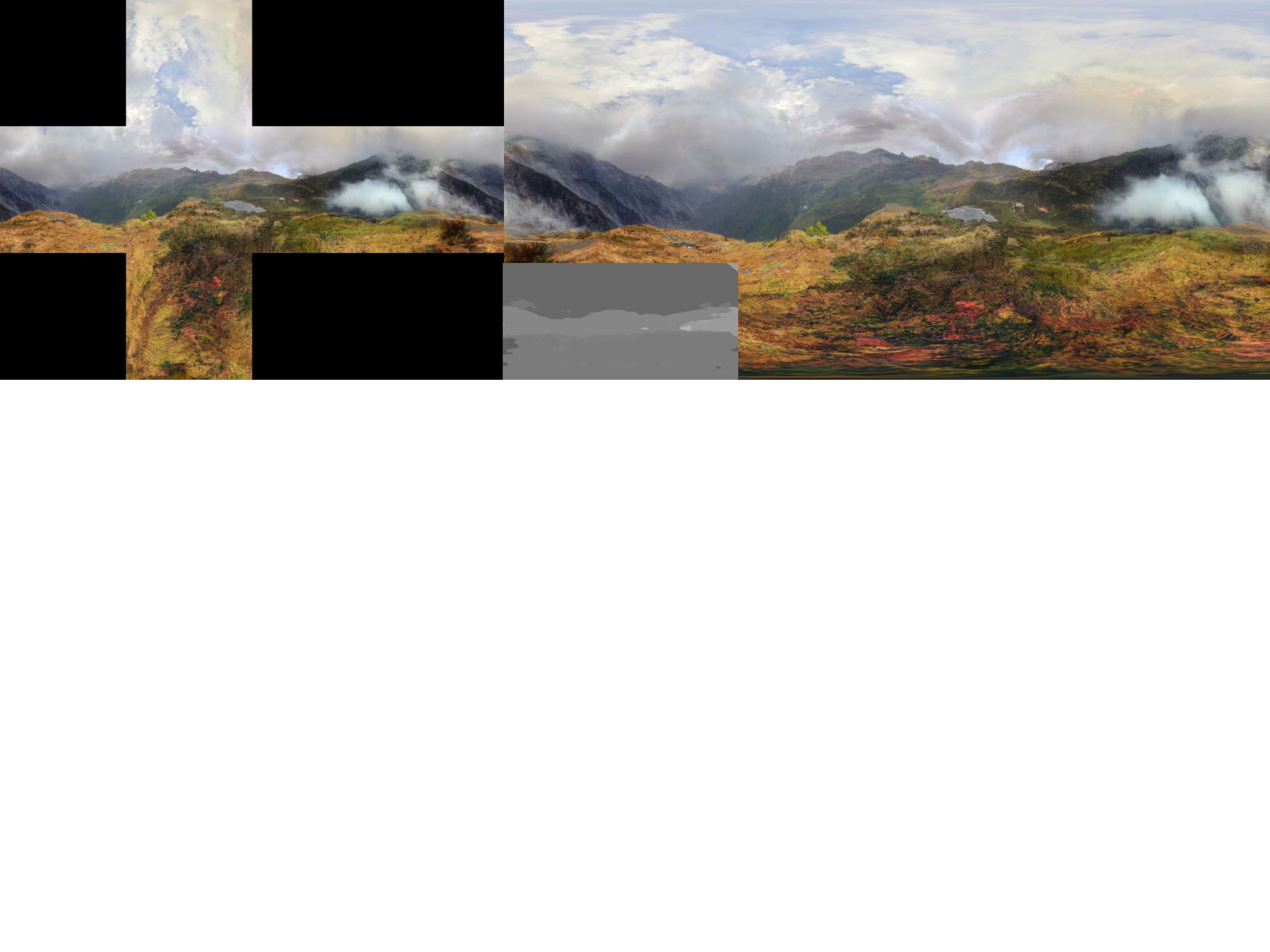}
    \caption{
        \ours on 360 cubemap image. (Left) Cubemap representation; (Right) Equirectangular representation; (Lower middle) Semantic segmentation map used to condition the model.
    }
\end{figure*}

\end{document}